\documentclass{article}

\usepackage{microtype}
\usepackage{graphicx}
\usepackage{booktabs} 
\usepackage{amsmath}
\usepackage{multirow}
\usepackage{amsfonts}
\usepackage{subcaption}

\usepackage{hyperref}

\usepackage[accepted]{icml2025}

\icmltitlerunning{BackSlash: Rate Constrained Optimized Training of Large Language Models}

\begin{document}

\twocolumn[
\icmltitle{BackSlash: Rate Constrained Optimized Training of Large Language Models}

\icmlsetsymbol{equal}{*}

\begin{icmlauthorlist}
\icmlauthor{Jun Wu}{tsinghua}
\icmlauthor{Jiangtao Wen}{newyork}
\icmlauthor{Yuxing Han}{tsinghua}
\end{icmlauthorlist}

\icmlaffiliation{tsinghua}{Shenzhen International Graduate School, Tsinghua University}
\icmlaffiliation{newyork}{Computer Science, New York University}

\icmlcorrespondingauthor{Yuxing Han}{yuxinghan@sz.tsinghua.edu.cn}
\icmlcorrespondingauthor{Jiangtao Wen (project lead)}{jw9263@nyu.edu}

\icmlkeywords{Model Compression, Rate-Distortion Optimization, Entropy Encoding}

\vskip 0.3in
]

\printAffiliationsAndNotice{} 

\begin{abstract}
The rapid advancement of large-language models (LLMs) has driven extensive research into parameter compression after training has been completed, yet compression during the training phase remains largely unexplored. In this work, we introduce Rate-Constrained Training (BackSlash), a novel training-time compression approach based on rate-distortion optimization (RDO). BackSlash enables a flexible trade-off between model accuracy and complexity, significantly reducing parameter redundancy while preserving performance. Experiments in various architectures and tasks demonstrate that BackSlash can reduce memory usage by 60\% - 90\% without accuracy loss and provides significant compression gain compared to compression after training. Moreover, BackSlash proves to be highly versatile: it enhances generalization with small Lagrange multipliers, improves model robustness to pruning (maintaining accuracy even at 80\% pruning rates), and enables network simplification for accelerated inference on edge devices.
\end{abstract}

\section{Introduction}
As the foundation of modern artificial intelligence, generative large language models (LLMs) such as Llama \citep{Touvron2023LLaMAOA}, GPT \citep{Brown2020LanguageMA}, and Qwen \citep{Bai2023QwenVLAV} exhibit remarkable self-learning and non-linear modeling capabilities. With continuous advancements in deep learning, the parameter scales of LLMs have grown at an unprecedented rate, as shown in Table~\ref{table:gpt}. Looking ahead, it is expected that the parameter scale of neural networks will continue to expand rapidly, driving further progress in AI development.


\begin{table}[t]
\caption{Parameter scale and growth rate of GPTs as an example over recent years.}
\label{table:gpt}
\vskip 0.1in
\begin{center}
\begin{small}
\begin{tabular}{cccc}
\toprule
\textbf{Model name} & \textbf{Time} & \textbf{Parameter size} & \textbf{Growth rate}\\
\midrule
GPT-1 & 2018.06 & 117M & - \\
GPT-2 & 2019.02 & 1.5B & 12.8x \\
GPT-3 & 2020.06 & 175B & 116.7x \\
GPT-4 & 2024.11 & 1.8T & 1200.0x \\
\bottomrule
\end{tabular}
\end{small}
\end{center}
\vskip 0.1in
\end{table}

The ever-increasing size of LLM parameters leads to increased computational costs, inference latency, and network distribution overhead during model deployment. To enable efficient inference of LLMs on edge devices, extensive research has focused on model compression using techniques such as parameter quantization, model pruning, and low-rank matrix decomposition. However, while there have been extensive studies on LLM redundancy at the microscopic level, such as precision and structural inefficiencies, the overall parameter distribution has received little attention. In addition, most existing compression techniques are applied after training, as opposed to being integrated into the LLM parameter training process to proactively achieve optimized trade-offs between parameter precision, model size, and model performance. Finally, existing studies assume that model parameters follow the Gaussian distribution and, therefore employ Huffman coding designed using empirical statistics for compression. Both the probabilistic model and the entropy coding technique have room for improvement.

In this paper, we introduce a rate-constrained optimized approach to LLM training. By incorporating model parameter size into the training process through a rate ($R$) and distortion ($D$) joint optimization, the proposed rate-constrained training (BackSlash) approach is capable of producing the optimal performing model for a given parameter set size, that is, producing the best-fit model given by its end application(hardware constrained parameter set size).



The main contributions of this paper are as follows:
\begin{enumerate}
    \item Instead of the widely used Gaussian model for LLM parameter distribution, we found through extensive experiments that the generalized Gaussian (GG) distribution with the shape parameter less than 2 is a better model.
    \item We propose to use exp-Golomb (EG) codes for entropy coding of LLM parameters, whose distribution can be well-modeled by GG distributions. It has been shown \cite{761289} that for GG sources, EG codes can achieve coding efficiency very close to the entropy limit, well over 90\% in many cases. We also find the optimal EG code with k=0 implementation can accommodate many applications. 
    \item Based on the GG distribution observation and using EG codes as entropy codes, we proposed a discretized generalized Gaussian information rate (DGGR) to measure the model information rate and an BackSlash algorithm that jointly optimizes the information rate and performance during the training phase of LLMs. Experiments with different LLMs and different deep-learning tasks show significant savings in model size as compared with both unconstrained training and unconstrained training followed by entropy coding.
\end{enumerate}

\section{Related Work}
\subsection{LLMs Compression}
To achieve low-cost distribution, deployment, and inference, many compression strategies for LLMs have been proposed.

Pruning reduces computational and storage overheads by removing unimportant weights or neurons from the model. Unstructured pruning achieves compression by removing redundant connections, e.g., \citet{Han2015LearningBW} and \citet{Han2015DeepCC} proposed a pruning method based on weight paradigms. Because unstructured pruning may lead to irregular network structures, structured pruning of filters or channels was proposed. \citet{Li2016PruningFF} proposed pruning based on filter, while \citet{Luo2017ThiNetAF} proposed Thinet that minimized the reconstruction error. Hardware constraints \cite{He2018AMCAF, Wang2018HAQHA} such as energy consumption and delay were also introduced into the pruning process to optimize the model performance in resource-constrained environments. Prune continues to be an important direction for model optimization, as evidenced by recent publications such as Dynamic Structure Pruning \citep{Park2023DynamicSP}, LAPP \citep{Zhai2023LAPPLA}, and Turbo-VBI \citep{Xia2023StructuredBC}.

Quantization can speed up training and inference by reducing the precision of weights and activation values. Binary weights \citep{Courbariaux2015BinaryConnectTD, Rastegari2016XNORNetIC}, triple weights \cite{Li2016TernaryWN, Zhu2016TrainedTQ}, cluster quantization \cite{Gong2014CompressingDC, Choi2016TowardsTL}, and mixed bit-width quantization \cite{Zhou2016DoReFaNetTL, 8578558} were examples of quantization techniques. \citet{long2020novel} used shift operation to replace the costly full-precision operation by quantizing low-bit weights and activations. \citet{Liu2021NonuniformtoUniformQT} simultaneously maintained the representational power of non-uniform quantization and the efficiency of uniform quantization.

Low-rank decomposition \cite{jaderberg2014speeding, Masana2017DomainAdaptiveDN}, parameter sharing \cite{Wang2017BeyondFC, Kossaifi2019TNetPF}, and knowledge distillation \cite{Xu2018PADNetMG, Chen2017DarkRankAD} have demonstrated significant effectiveness in applications. Additionally, with large-scale distributed deep learning training systems, communication-efficient gradient compression techniques were proposed \cite{Lin2017DeepGC}. These approaches collectively enhance the efficiency of deep learning model training and deployment.

\subsection{Rate Distortion Optimization}
Information theory \cite{shannon1948mathematical, cover1999elements} mathematically quantified the efficiency with which information can be transmitted, stored, and processed, where rate-distortion function defines the minimal distortion that can be achieved while entropy coding a system to a given bitrate \cite{davisson1972rate, berger2003rate}.

In practical applications, rate-distortion optimization (RDO) has found extensive adoption in video coding \cite{luttrell2000trellis, Jtc2010AdvancedVC, wien2015high, brand2022rdonet, chen2023rate, guo2023pre, chiang2023multi, xia2023gan, zhang2024efficient}, etc. are also continuing to deepen the application of RDO in video and images.

In recent years, RDO has also been introduced for the compression of neural networks. For example, \citet{Gao2018RateDF} investigated the fundamental limits of model compression and proposed a compression framework for pruning, quantization, and other techniques. \citet{Isik2021SuccessivePF} proposes a new pruning strategy based on RDO to approach the compression limits of neural networks. In both cases, RDO is applied after models have been trained to further pruning and quantization, as opposed to being integral to the training process itself.

\section {Generalized Gaussian Model of LLM Parameters} \label{sec:gg}
Most research assumes that LLM parameters follow the Gaussian distribution in the initialization and rarely discussed the distribution after training. For example, Gaussian distribution was used by both Xavier and He for random parameter initialization \cite{glorot2010understanding, He2015DelvingDI}. However, through extensive experiments, we found that the more broad generalized Gaussian (GG) distribution family with shape parameter less than 2 might be a better model for LLM models, especially considering that different regulations during training may impact parameter distribution. The distribution usually also changes during training as the model converges. In practice, the parameter distribution tends to develop heavier tails during the training process\cite{Fortuin2021BayesianNN}.

Mathematically, the probability density function (pdf) of generalized Gaussian distribution is defined as 
\begin{equation}
    \label{eq:pdf}
    f(x)=C_1e^{-C_2|x|^{\nu}}
\end{equation}
where 
\begin{eqnarray}
C_1=\frac{\nu\gamma}{2\Gamma(1/\nu)}, C_2=\gamma^{\nu}, \\ 
\gamma=\frac{1}{\sigma}\sqrt{\frac{\Gamma(3/\gamma)}{\Gamma(1/\gamma)}},\nonumber\\
\Gamma(\alpha)=\int_{0}^{\infty}t^{\alpha-1}e^{-t}dt, \nonumber
\end{eqnarray}
and 
$\alpha>0$.

It is easy to see that when $\nu=1$, the GG distribution is the Laplacian distribution, while when $\nu=2$, it is the Gaussian distribution. Varying the shape parameter of GG distribution allows for better match between the probabilistic model to better match LLM while using the same mathematical formulation.

The GG distribution in (\ref{eq:pdf}) is a continuous distribution, while in reality, LLM parameters all have fixed-length and limited precision. Therefore, we treat LLM parameters as a quantized GG distribution. Assuming the quantization step size of the parameters is $\delta$, the probability of a parameter $\theta_i$ is
\begin{equation}
p(\theta_i)=\int_{\theta_i}^{\theta_i+\delta}f(x)dx.    
\end{equation}
As $\delta$ is typically small, we approximate 
\begin{equation}
p(\theta_i)\approx\delta f(\theta_i)=\delta C_1 e^{-C_2|\theta_i|^{\nu}}.
\end{equation}

\begin{figure*}[t]
    \centering
    \begin{subfigure}{0.23\textwidth}
        \includegraphics[width=\linewidth]{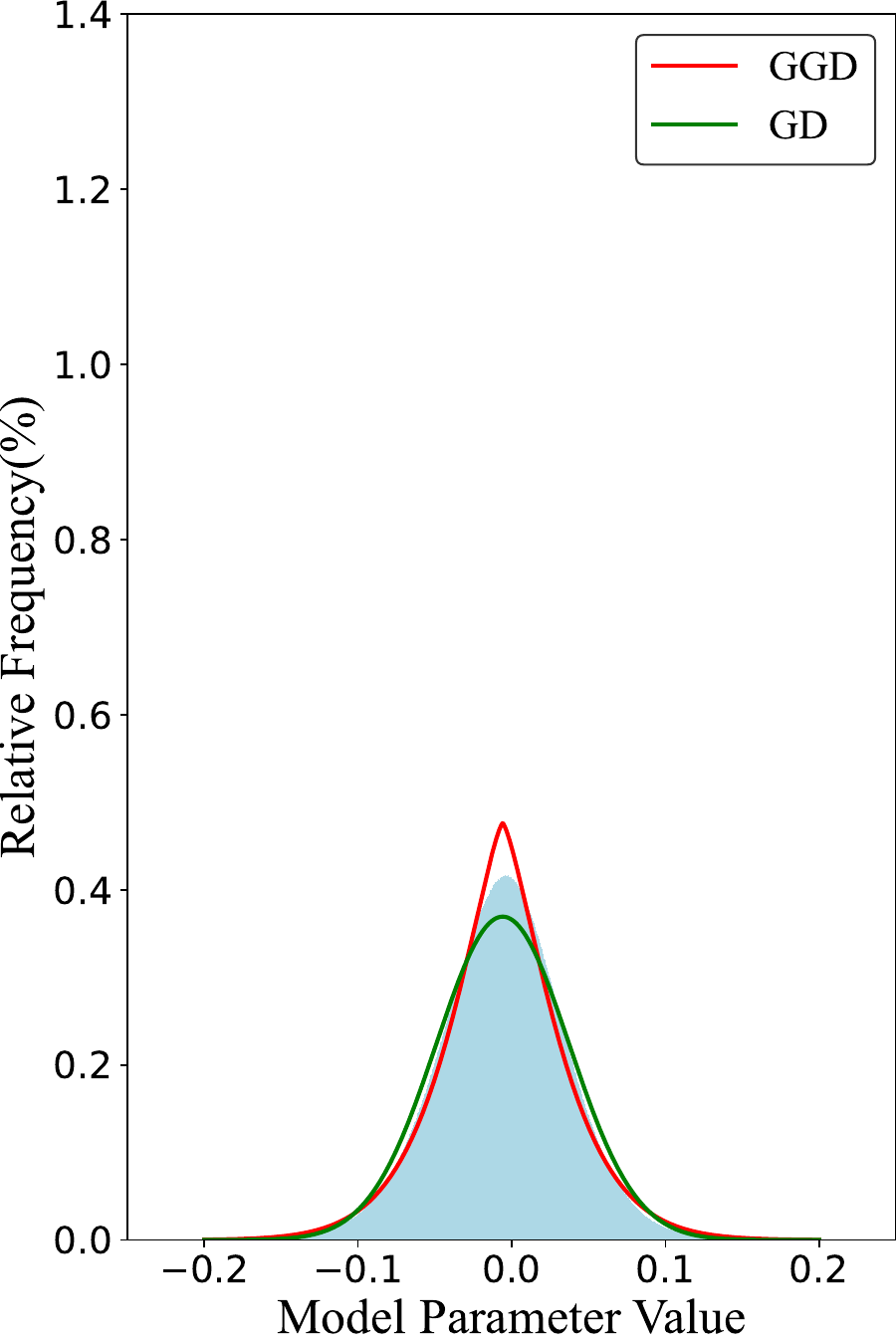}
        \caption{BERT}
    \end{subfigure}
    \hspace{0.05in}
    \begin{subfigure}{0.23\textwidth}
        \includegraphics[width=\linewidth]{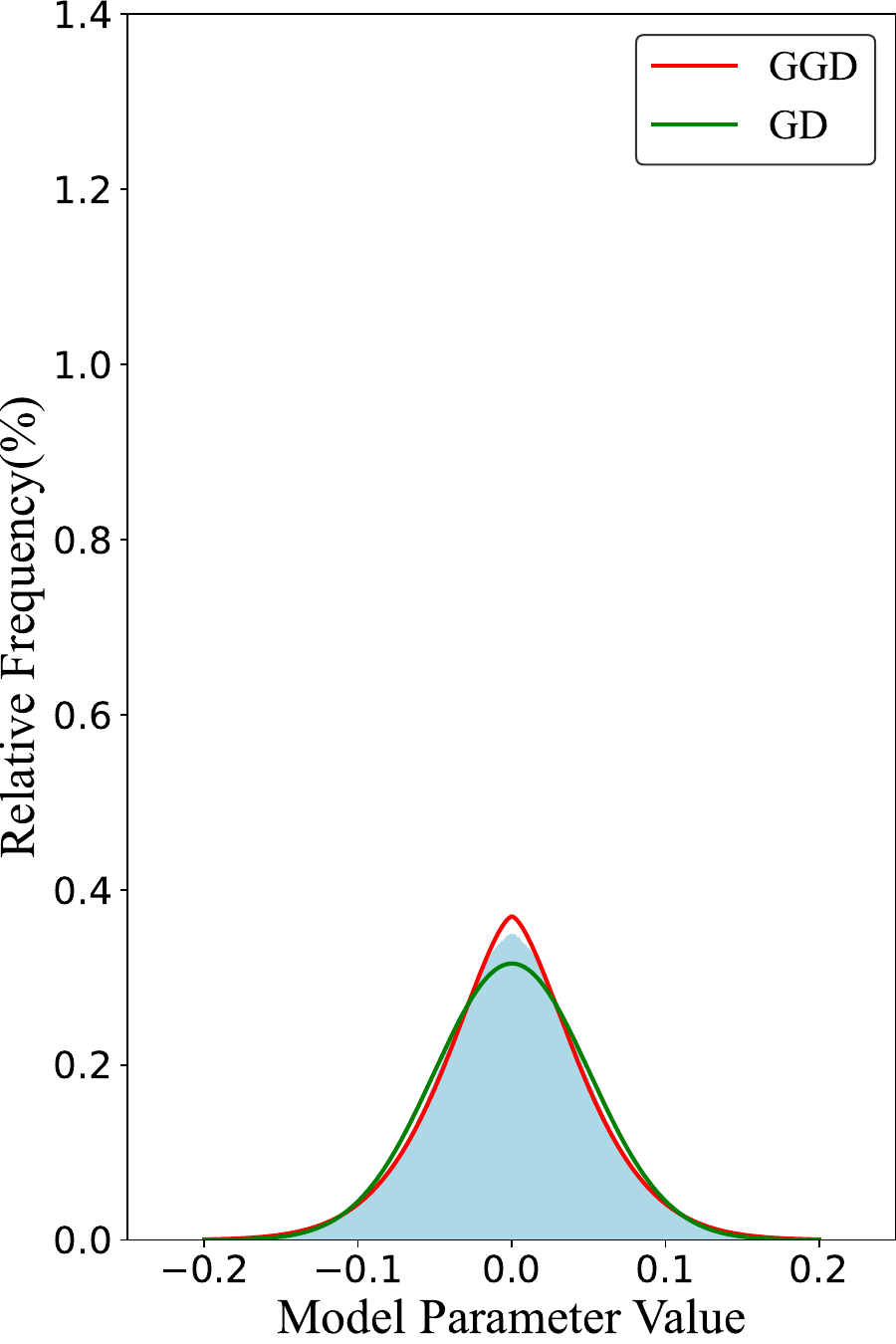}
        \caption{GPT2}
    \end{subfigure}
    \hspace{0.05in}
    \begin{subfigure}{0.23\textwidth}
        \includegraphics[width=\linewidth]{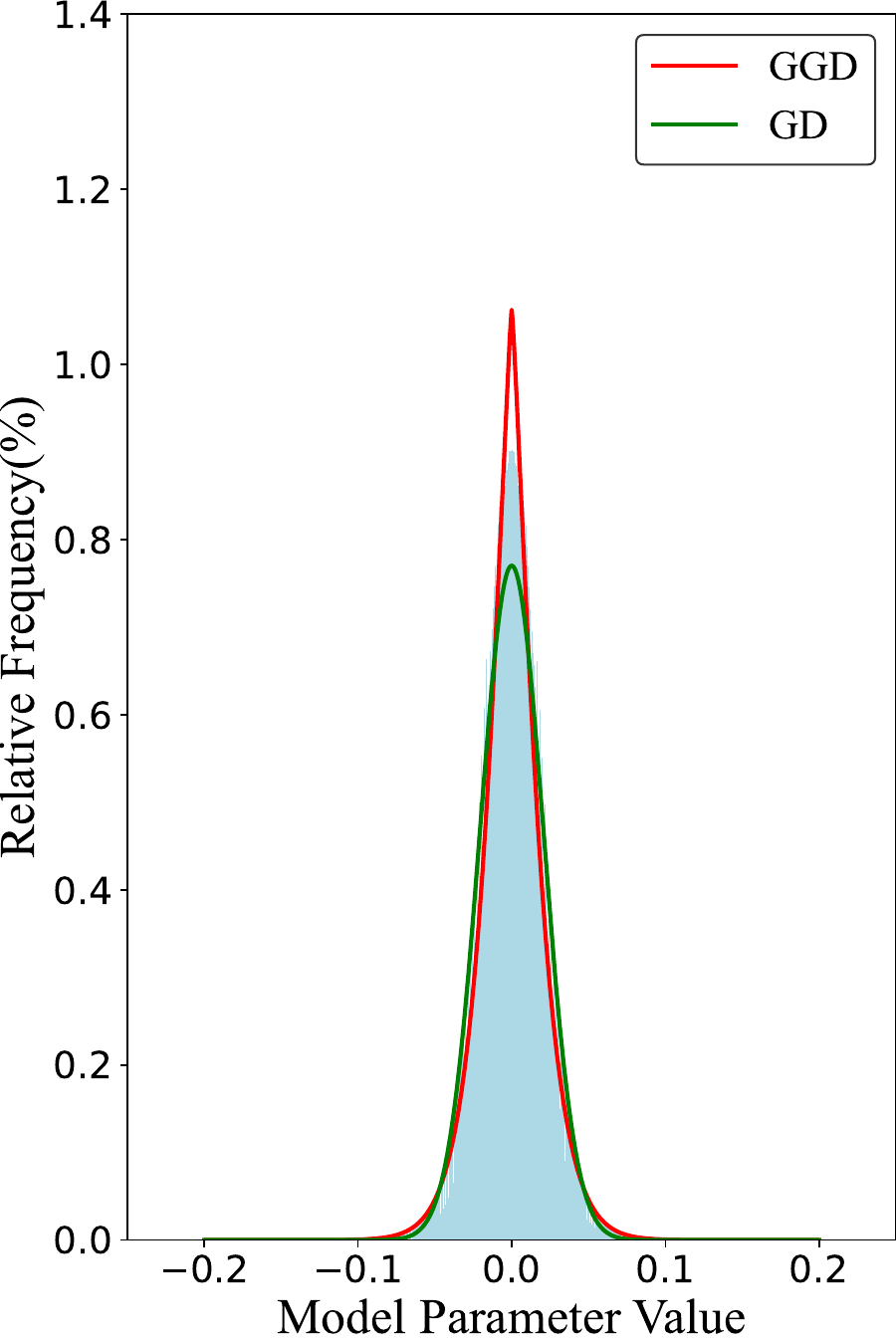}
        \caption{Llama3}
    \end{subfigure}
    \hspace{0.05in}
    \begin{subfigure}{0.23\textwidth}
        \includegraphics[width=\linewidth]{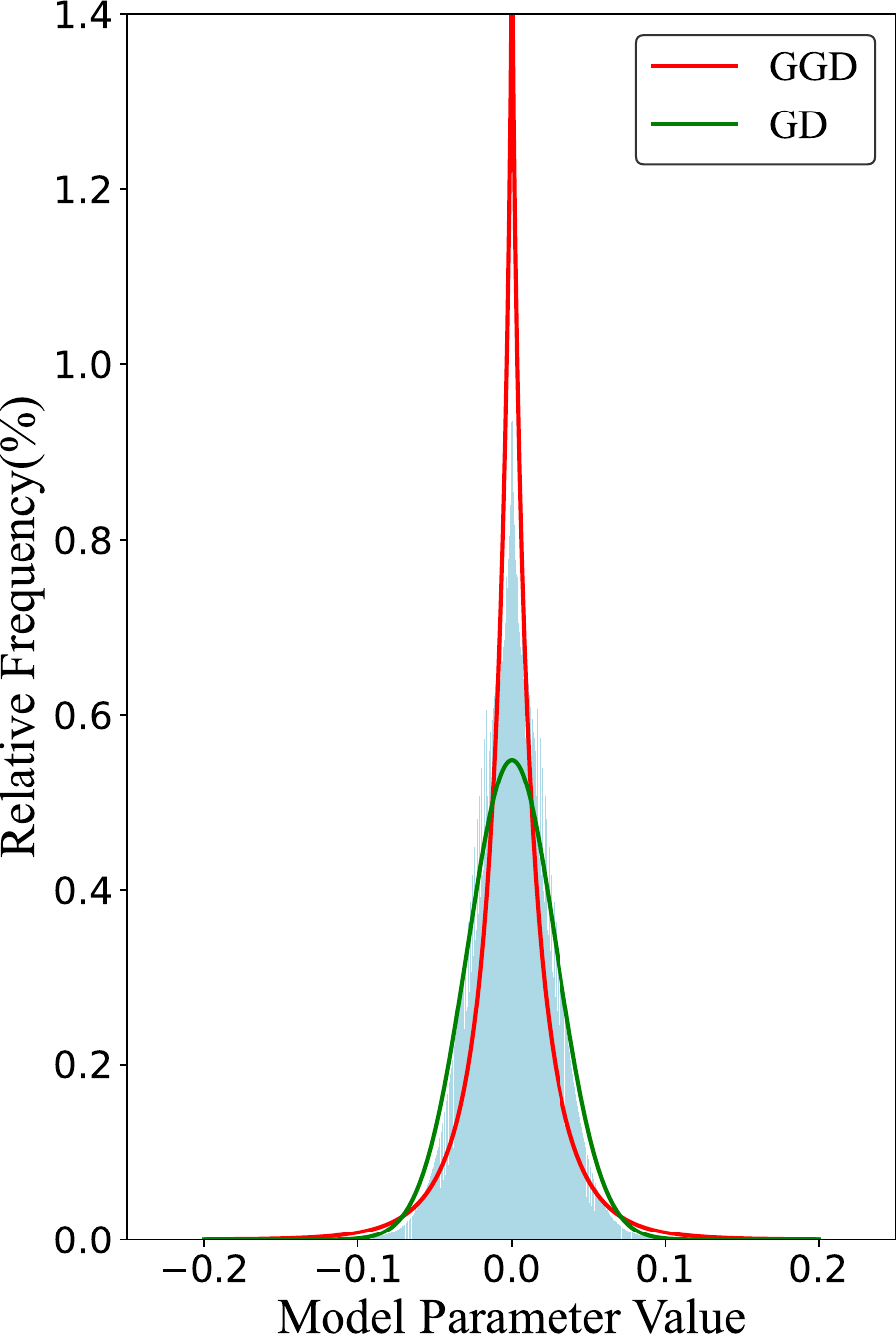}
        \caption{DeepSeek}
    \end{subfigure}
    \caption{Parameter distributions fitting by generalized Gaussian distribution (GGD) and Gaussian distribution (GD) under different LLMs. GGD fits the boundaries of the parameter distributions better than GD does.}
    \vspace{0.2 in}
    \label{fig:model-dis}
\end{figure*}

The validity of this assumption could be verified with existing LLMs. For instance, BERT-base (110M) can be well-modeled by a GG with a shape parameter of 1.36, the shape parameter for GPT2 (774M) is 1.54, or 1.26 for Llama3 (1B), or 0.85 for DeepSeek (7B), and their distributions are shown in Fig.~\ref{fig:model-dis}. These shape parameter values are, although different, all smaller than 2. The corresponding pdfs show higher peaks and longer tails than the Gaussian distribution.

Denote the size of LLM as $N_p$, the mean of the information content of the parameters can be calculated as $R(\theta)=-\frac{1}{N_p}\sum_{i=1}^{N_p}\log_2 p(\theta_i)$. Neglecting constant factors and terms, we define discretized generalized Gaussian rate (DDGR) as follows
\begin{equation}
    \label{eq:dggr}
    R(\theta)=\frac{1}{N_p}\sum_{i=1}^{N_p}|\theta_i|^{\nu},
\end{equation}
and use $R(\theta)$ as a measure of the information complexity of the model.

\section{Rate-Constrained Training (BackSlash) of LLMs}
\subsection{Overview and loss function definition}
In contrast to traditional non-constrained training, the target loss function for optimization in BackSlash considers both model performance and model size is
\begin{equation}
    \mathcal{J} = D + \lambda \cdot R,
    \label{eq:rdo}
\end{equation}
where $D$ denotes the distortion of the fitting data, i.e. the deviation between the model predictions and the ground truth. $R$ denotes the rate of the model parameters, indicating the complexity of the model itself. $\lambda$ is the Lagrange multiplier. The selection of distortion $D$ varies depending on the deep-learning task. For example, we often use the categorical cross-entropy loss function in classification problems and the mean squared error loss in regression tasks. Methods like KL divergence or logarithmic loss are also utilized in some specific tasks. We can choose the most suitable empirical loss function for specific tasks, which does not affect the BackSlash results.

The rate $R$ is expressed by the average information content of the parameters, defined using DGGR. Combining (\ref{eq:dggr}) and (\ref{eq:rdo}) we get
\begin{equation}
    \label{eq:rd-cost}
    \mathcal{J}=\mathcal{L}(X,Y,f,\theta)+\lambda\cdot \frac{1}{N_p}\sum_{i=1}^{N_p}|\theta_i|^{\nu},
\end{equation}
where $X$ and $Y$ denote the training ground truth, $f$ and $\theta$ denote the forward propagation function and parameter set of the neural network.

The shape parameter $\nu$ of DGGR is not a constant during training and needs to be dynamically estimated before each batch of gradient descent. A well-known method \cite{Sharifi1995EstimationOS} for estimating the shape parameter is by introducing a comparison function:
\begin{equation}
    \label{eq:shape-est}
    \rho(\nu) = \frac{\Gamma(1/\nu) \cdot \Gamma(3/\nu)}{\Gamma^2(2/\nu)} = \frac{\mathbb{E}[\theta^2]}{\mathbb{E}^2[|\theta|]}.
\end{equation}

Specifically, the estimation process of $\nu$ can be organized as follows: 
\begin{enumerate}
    \item obtain the model parameters $\theta$ and compute $\mathbb{E}[\theta^2]$ and $\mathbb{E}^2[|\theta|]$,
    \item estimate $\rho(\nu)$ by (\ref{eq:shape-est}) based on $\mathbb{E}[\theta^2]$ and $\mathbb{E}^2[|\theta|]$,
    \item Find $\nu$ using $\rho(\nu)$.
\end{enumerate}

Additionally, notice that when $ 0 < \nu < 1 $ and $ \theta \to 0 $, $ \nabla R(\theta) \to \infty $. This causes severe oscillations in the parameters during gradient descent, which prevents the model from converging. To address this, we introduce a trick to optimize the gradient descent of $ R(\theta) $ by adding a constant $\epsilon$ ($\epsilon>0$) to control the gradient size and modify the information rate formula as ${R}(\theta) = \frac{1}{N_p} \sum_{i=1}^{N_p} (|\theta_i| + \epsilon)^{\nu}$. We refer to this method of gradient suppression as \textbf{soft gradient clipping}.

\subsection{BackSlash algorithm description}
\label{sec:algo}
The overall BackSlash algorithmic can be summarized as follows:
\renewcommand{\thealgorithm}{1} 
\begin{algorithm}
    \caption{Rate-Constrained Training (BackSlash)}
    \begin{algorithmic}[1]
        \STATE \textbf{Require:} Model $f$, learning rate $\eta$, loss function $\mathcal{L}$, Lagrange multiplier $\lambda$, and clipping coefficient $\epsilon$.
        \FOR{each epoch $\tau$}
            \FOR{each batch $(x_i, y_i)$}
                \STATE Retrieve all model parameters $\theta$.
                \STATE Estimate comparison function $\rho(\nu)$: $\rho(\nu)\leftarrow\frac{\mathbb{E}[\theta^2]}{\mathbb{E}^2[|\theta|]}$
                \STATE Find shape parameter $\nu$ using $\rho(\nu)$.
                \STATE Forward propagation and calculate RD Cost $\mathcal{J}$: $\mathcal{J}\leftarrow\mathcal{L}(x_i, y_i, f, \theta) + \lambda \cdot \frac{1}{N_p} \sum_{i=1}^{N_p} (|\theta_i| + \epsilon)^{\nu}$
                \STATE Backward propagation and optimize parameters: $\theta\leftarrow\theta-\eta\cdot(\frac{\partial\mathcal{L}}{\partial\theta}+\lambda\cdot\frac{\nu\theta}{N_p|\theta|}(|\theta|+\epsilon)^{\nu-1})$.
            \ENDFOR
        \ENDFOR
        \STATE \textbf{Until} convergence or max iterations.
    \end{algorithmic}
\end{algorithm}

It should be noted that if we set the shape parameter $\nu$ in DGGR to be $\nu=1$ and $\nu=2$, we find that DGGR degenerates into $L_1$ \cite{9742936} regularization ($\frac{1}{N_p}\sum_{i=1}^{N_p}|\theta_i|$) and $L_2$ \cite{hoerl1970ridge} regularization ($\frac{1}{N_p}\sum_{i=1}^{N_p}|\theta_i|^{2}$) respectively. This means, that $L_1$ and $L_2$ regularization are special cases of DGGR when the model parameter distribution is the Laplace and the Gaussian distributions respectively.

\subsection{Entropy coding of LLM using EG codes}

\begin{table*}[ht]
\caption{The Structure of exp-Golomb code with different parameter $k$ which is from 0 to 5 as an example. In general, EG codes with a smaller parameter k encode better for GG sources with low shape parameters.}
\label{table:eg-code}
\vskip 0.1in
\begin{center}
\begin{small}
\begin{tabular}{c|ccccccccccc}
\toprule
Parameter ($k$) & \textbf{0} & \textbf{1} & \textbf{2} & \textbf{3} & \textbf{4} & \textbf{5} & \textbf{6} & \textbf{7} & \textbf{8} & \textbf{9} & $\cdots$ \\
\midrule
$k=0$ & 1 & 010 & 011 & 00100 & 00101 & 00110 & 00111 & 0001000 & 0001001 & 0001010 & $\cdots$ \\
$k=1$ & 10 & 11 & 0100 & 0101 & 0110 & 0111 & 001000 & 001001 & 001010 & 001011 & $\cdots$ \\
$k=2$ & 100 & 101 & 110 & 111 & 01000 & 01001 & 01010 & 01011 & 01100 & 01101 & $\cdots$ \\
$k=3$ & 1000 & 1001 & 1010 & 1011 & 1100 & 1101 & 1110 & 1111 & 010000 & 010001 & $\cdots$ \\
$k=4$ & 10000 & 10001 & 10010 & 10011 & 10100 & 10101 & 10110 & 10111 & 11000 & 11001 & $\cdots$ \\
$k=5$ & 100000 & 100001 & 100010 & 100011 & 100100 & 100101 & 100110 & 100111 & 101000 & 101001 & $\cdots$ \\
\bottomrule
\end{tabular}
\end{small}
\end{center}
\vskip 0.3in
\end{table*}

The complexity constraint in BackSlash is defined using DGGR. In practice, the parameters will have to be entropy coded using practical entropy codes, whose rate can not exactly match the DGGR.

Due to its simplicity, Huffman codes have been used for entropy coding of LLM parameters. For example, in \citet{Han2015DeepCC} achieved 20\%-30\% size reduction using Huffman coding. However, using Huffman coding in BackSlash or compression of LLM in general has several drawbacks. 

First of all, Huffman code tables are designed using explicit distributions calculated from LLM parameters. The mismatch between the distribution of the parameters and the distribution for which the Huffman code is designed may lead to severe coding efficiency loss. On the other hand, the large parameter size and non-parallelizable table building process of Huffman code may bring prohibitively high complexity to BackSlash. This is also why we used the theoretical DDGR as opposed to coded bits in the loss function.

Secondly, Huffman tables designed for different LLMs are different, while a practical implementation may often need to accommodate multiple models in the same system (e.g. on the same chip). 

Thirdly, the Huffman table designed based on empirical distributions usually is not well-structured, leading to more complicated encoder/decoder implementation. 

Fourthly, we observe the Huffman code can only provide minimal efficiency gains over EG code on BackSlash-trained models in all subsequent experiments.

As noted in Section \ref{sec:gg}, LLM parameters can be modeled well by quantized GG distributions with shape parameters less than 2. In \citet{761289}, Wen and Villasenor studied and proposed using exp-Golomb (EG) codes to entropy coding quantized GG sources. The structure of EG code is shown in Table~\ref{table:eg-code}. The advantages of EG codes can be summarized as the following:
\begin{enumerate}
    \item the efficiency of EG codes is consistently within a few percentage points of the entropy limit and almost identical to the Huffman code specifically designed for each quantized GG source,
    \item the performance of EG codes is robust with regard to parameter mismatch, and as a result, adaptive coding is not needed when parameters of the quantized GG source change,
    \item EG codes contain an infinite number of codewords, and can therefore be used for LLM of any size,
    \item EG codes are nicely structured, and allow for highly optimized encoder/decoder.
\end{enumerate}

Therefore, in our experiments, we used EG codes for the actual entropy coding rate (as opposed to the theoretical rate of DDGR in BackSlash) for both entropy coding of parameters after traditional, unconstrained LLM training, and in comparison, BackSlash. In our experiment, we tested EG with different parameters on several models with BackSlash which is shown in Table~\ref{table:model-eg}, and the EG code that we found was optimal was EG with parameter $k=0$.

\begin{table}[ht]
\caption{Average code lengths for several models with different EG parameters. With the EG parameter increasing, the average code length of the model parameters also increases and the EG code gradually converges to the fixed-length code.}
\label{table:model-eg}
\vskip 0.1in
\begin{center}
\begin{small}
\begin{tabular}{c|ccccccc}
\toprule
Model & \textbf{$k=0$} & \textbf{$k=1$} & \textbf{$k=2$} & \textbf{$k=3$} & \textbf{$k=4$} \\
\midrule
BERT & \textbf{2.64} & 3.16 & 3.74 & 4.42 & 5.17 \\
GPT-2 & \textbf{2.46} & 3.05 & 3.70 & 4.40 & 5.19 \\
Llama-3 & \textbf{1.72} & 2.47 & 3.26 & 4.12 & 5.03 \\
Gemma-2 & \textbf{1.16} & 2.10 & 3.06 & 4.02 & 5.01 \\
\bottomrule
\end{tabular}
\end{small}
\end{center}
\vskip 0.0in
\end{table}

In addition, we note that after BackSlash, most model parameters are zero, while the non-zero values are extremely sparse, usually accounting for few percent of all possible values. For example, the number of code words occupied by BERT with BackSlash after quantization with a step of $2^{-8}$ is 2695 but only 641 are actually used. And except for a small number of quantized parameters near 0, the others are highly disordered. For example, the quantized parameter $-177$ of BERT is only ranked 609 by frequency but actually it takes up 142nd. Therefore, prior to entropy coding of model parameters, we first rank the number of occurrences of all parameter values and map each value that model parameters might actually take to an index. The index, instead of the parameter value, is then entropy coded. This process can be formally summarized as follows:

\renewcommand{\thealgorithm}{2} 
\begin{algorithm}
    \caption{Parameter Entropy Encoding} 
    \begin{algorithmic}[1]
        \STATE \textbf{Require:} Model parameter set $\Theta$, quantization step size $2^{-n}$, encoding strategy $\text{Enc}$.
        \STATE Quantize the parameters: $Q \gets \text{round}(2^n \cdot \Theta)$
        \STATE Sort by frequency, build the code table by quantized parameter and sorted index: $\mathcal{C} \gets \{(q_i, c_i) \mid q_i \in Q_s, c_i \in C_s\}$
        \STATE Map quantized parameters to codewords and encode it to bitstream: $B_{stream} \gets \text{Enc}(\mathcal{C}[Q])$
        \STATE \textbf{Output:} Bitstream $B_{stream}$ and code table $\mathcal{C}$
    \end{algorithmic}
\end{algorithm}

\begin{figure*}[ht]
    \centering
    \begin{minipage}{0.45\textwidth}
        \includegraphics[width=\linewidth]{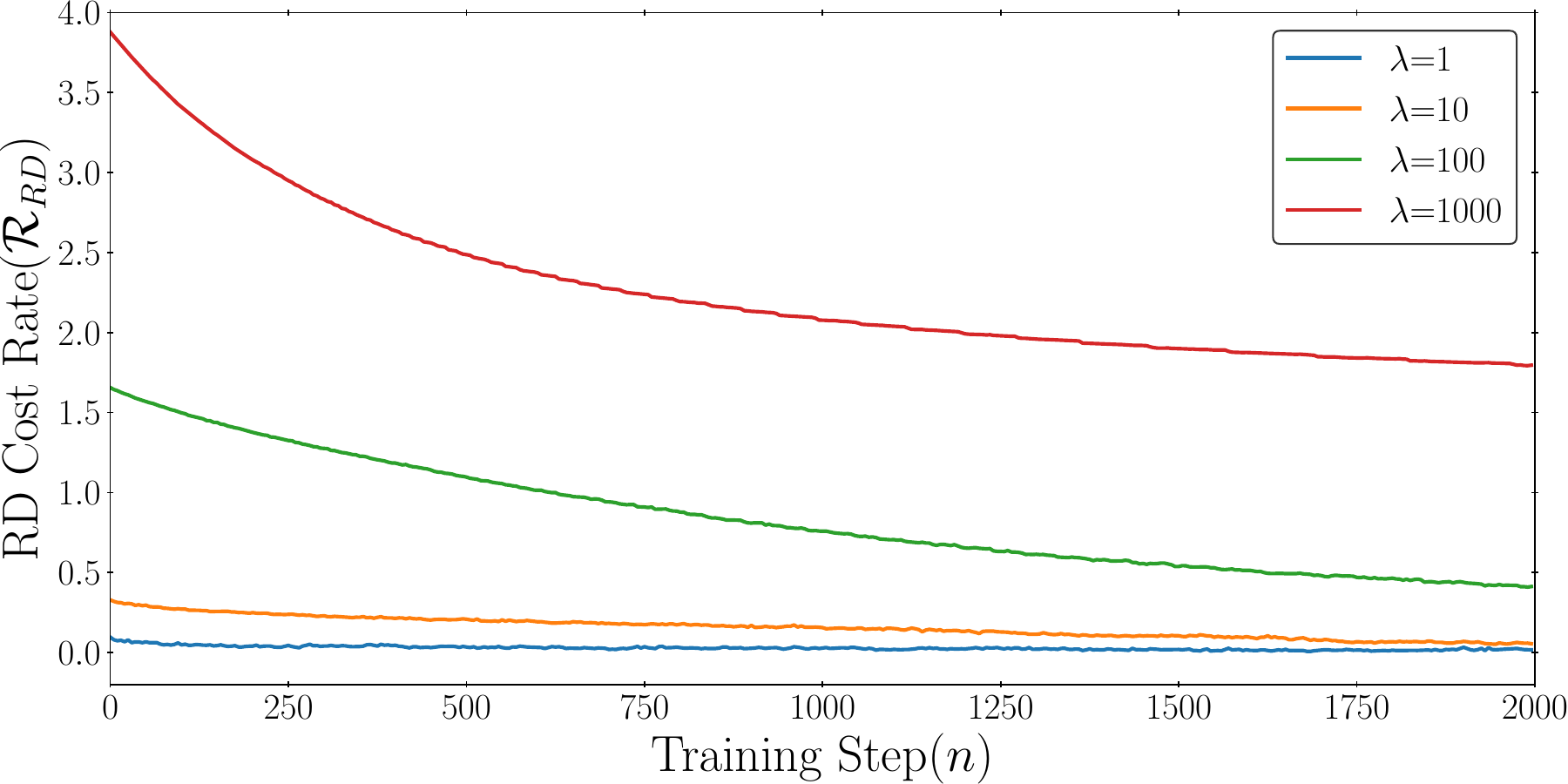}
        \vspace{-0.1in}
        \caption{RD cost rate changes in training with different Lagrange multiplier ($\lambda$).}
        \vspace{0.3in}
        \label{fig:loss}
    \end{minipage}
    \hspace{0.05in}
    \begin{minipage}{0.45\textwidth}
        \includegraphics[width=\linewidth]{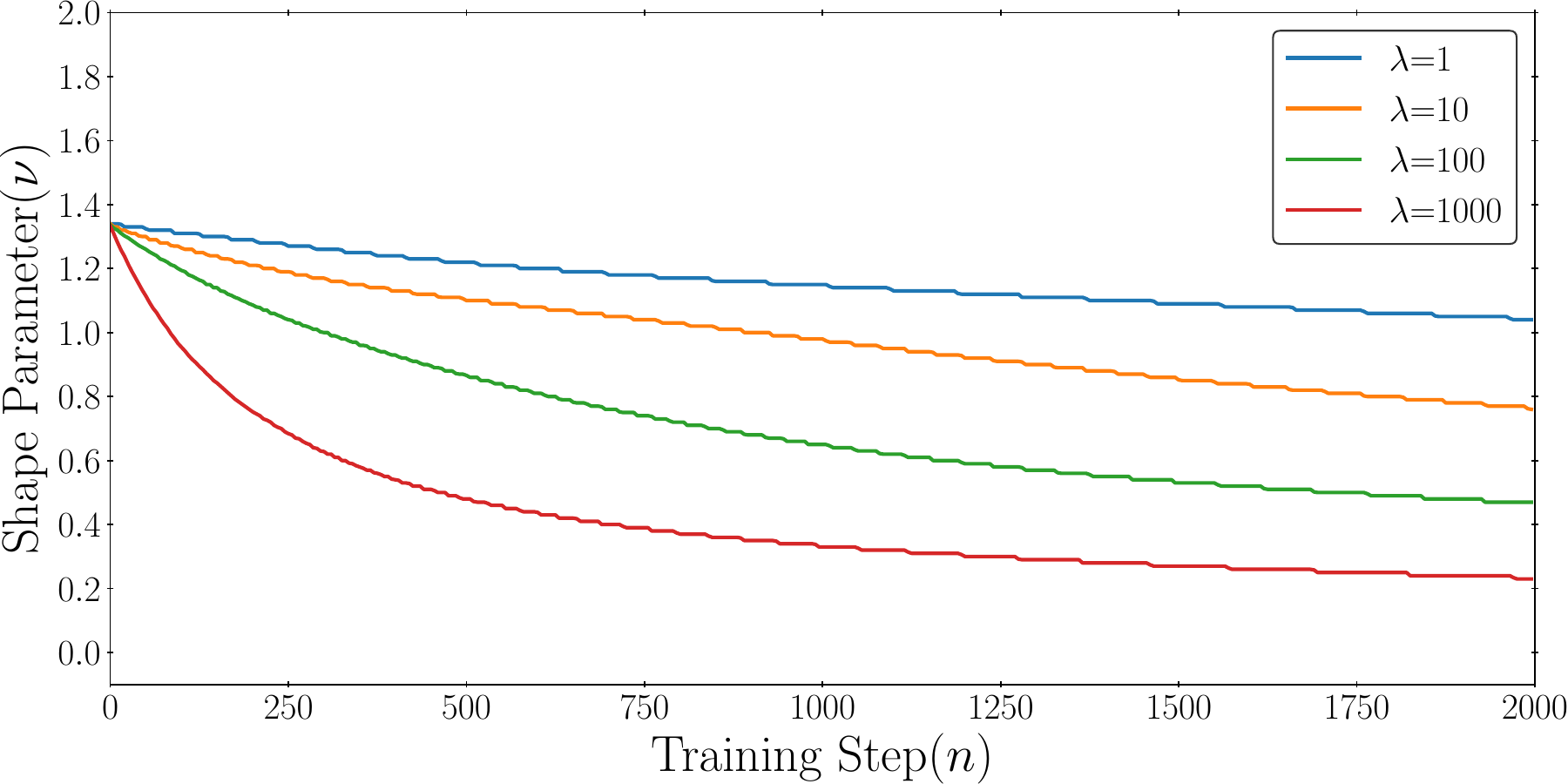}
        \vspace{-0.1in}
        \caption{Shape parameter changes in training with different Lagrange multiplier ($\lambda$).}
        \vspace{0.3in}
        \label{fig:shape}
    \end{minipage}
\end{figure*}
\begin{figure*}[ht]
    \centering
    \begin{minipage}{0.45\textwidth}
        \includegraphics[width=\linewidth]{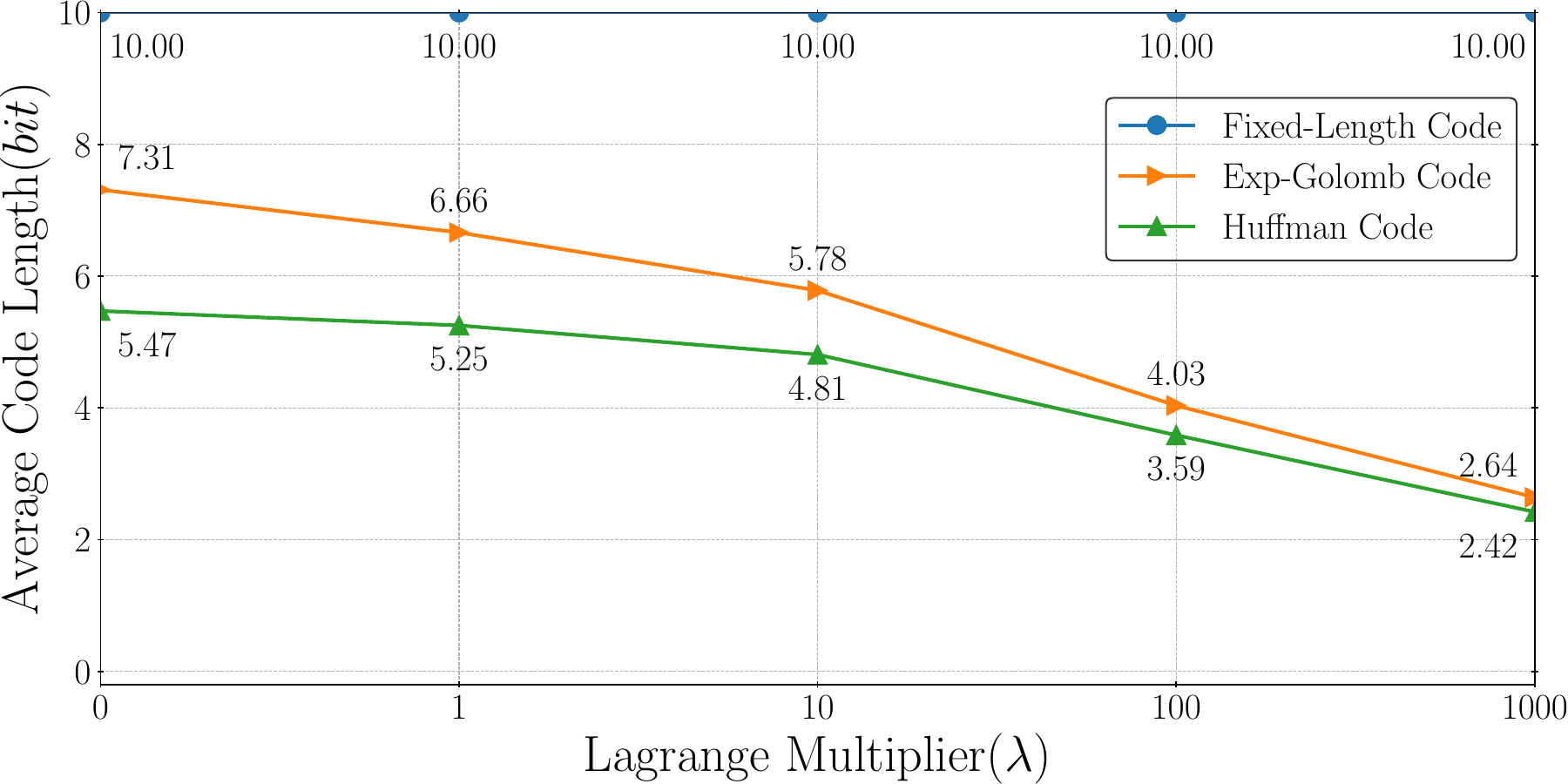}
        \vspace{-0.1in}
        \caption{Impact of Lagrange multipliers on average code length of various encoding algorithms.}
        \label{fig:avglength}
    \end{minipage}
    \vspace{0.1in}
    \hspace{0.05in}  
    \begin{minipage}{0.45\textwidth}
        \includegraphics[width=\linewidth]{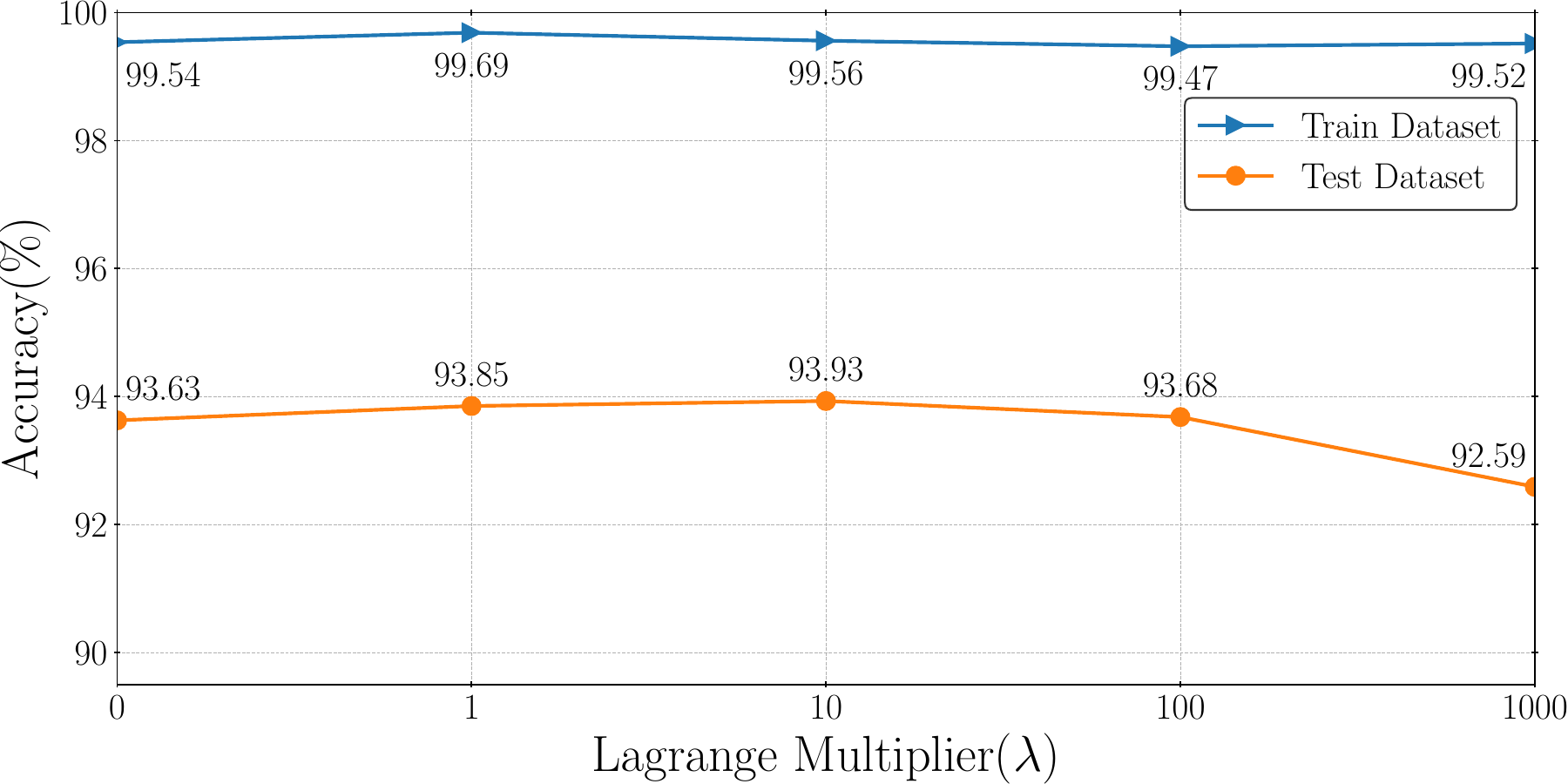}
        \vspace{-0.1in}
        \caption{Impact of Lagrange multipliers on accuracy of test and train Dataset.}
        \label{fig:accuracy}
    \end{minipage}
    \vspace{0.1 in}
\end{figure*}
\begin{figure*}[ht]
    \centering
    \begin{subfigure}{0.3\textwidth}
        \includegraphics[width=\linewidth]{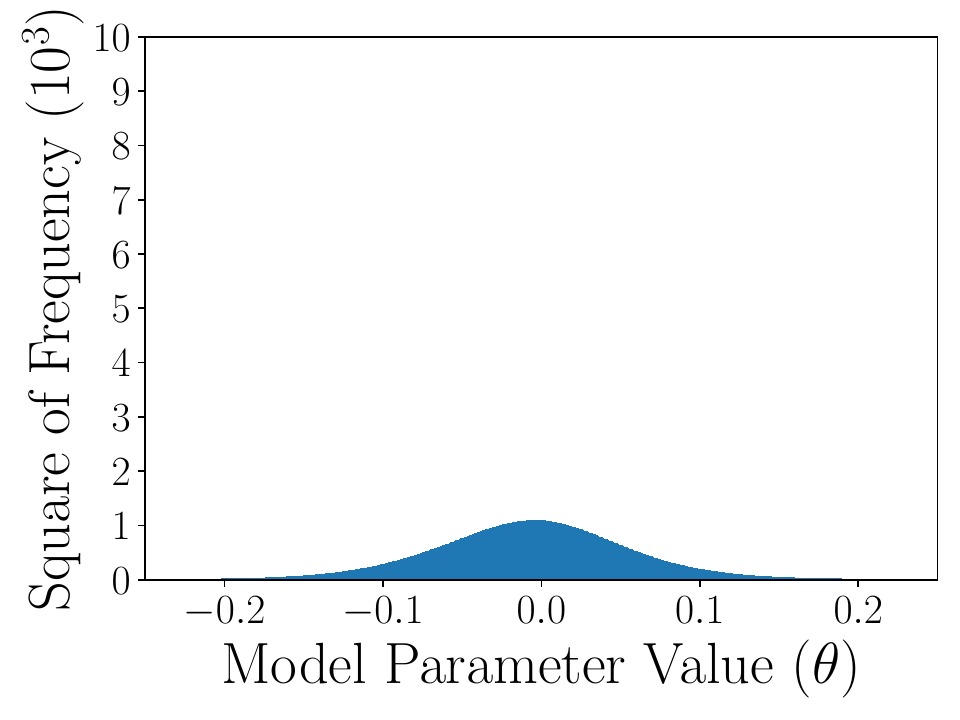}
        \caption{Lagrange Multiplier 0}
    \end{subfigure}
    \hspace{0.05in}
    \begin{subfigure}{0.3\textwidth}
        \includegraphics[width=\linewidth]{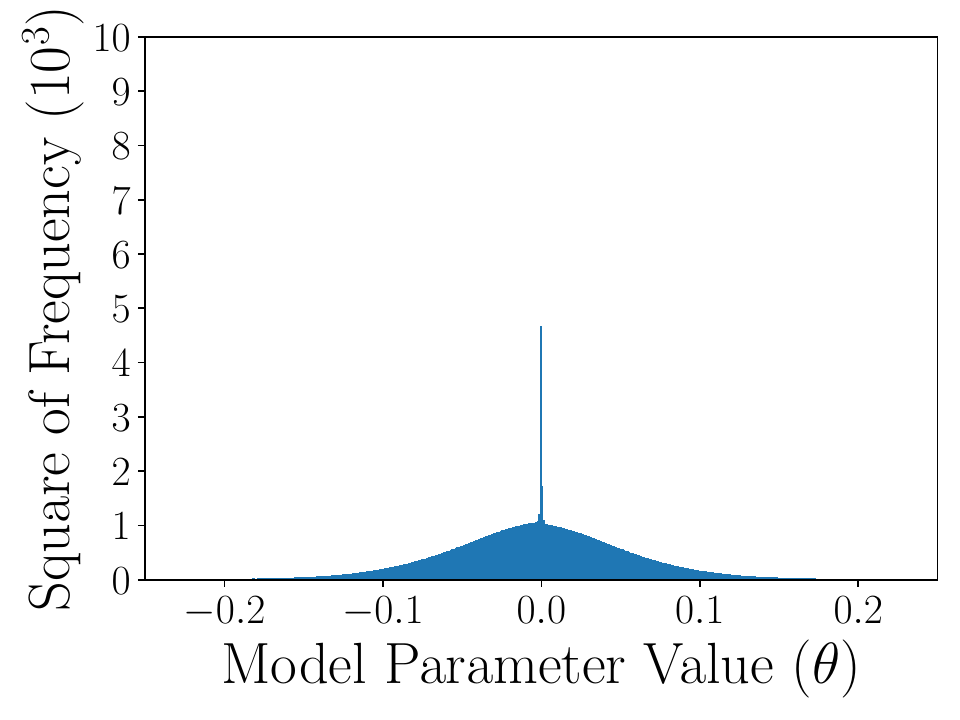}
        \caption{Lagrange Multiplier 10}
    \end{subfigure}
    \hspace{0.05in}
    \begin{subfigure}{0.3\textwidth}
        \includegraphics[width=\linewidth]{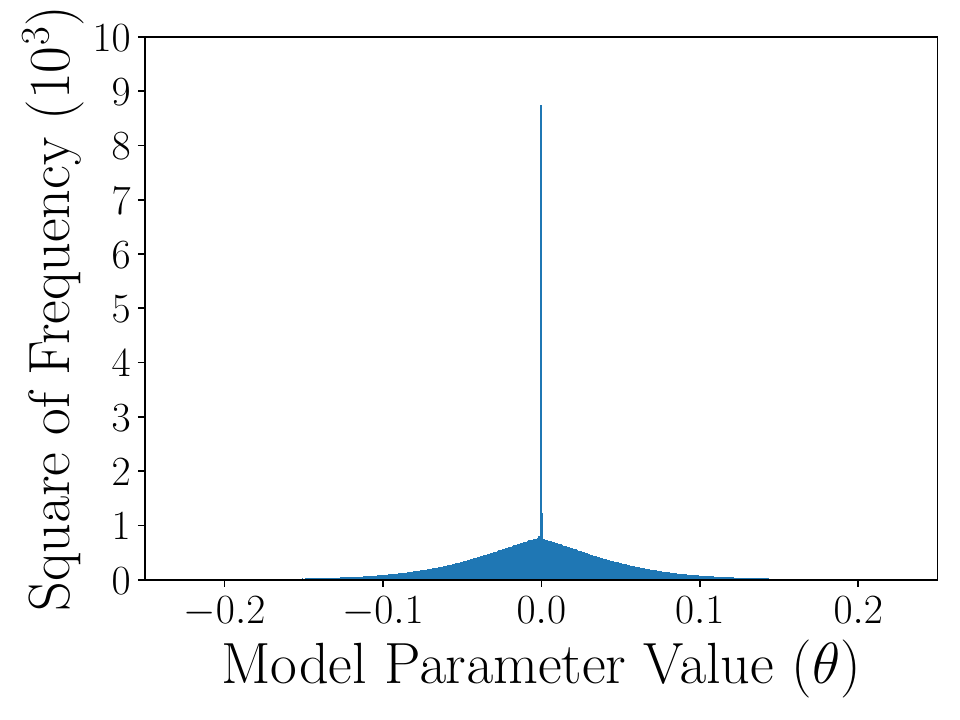}
        \caption{Lagrange Multiplier 1000}
    \end{subfigure}
    \caption{Parameters distribution under different Lagrange multiplier training. With the Lagrange multipliers increasing, the parameter distributions become more concentrated and have higher peaks and lower tails.}
    \vspace{0.2 in}
    \label{fig:distribution}
\end{figure*}

The distribution of indices still follows a generalized Gaussian, though the value mapping induces slight deviations compared to the original parameters. For example, the shape of parameters and index of BERT under normal training is 1.36 and 1.47, while under BackSlash they become 0.26 and 0.30, respectively. Nevertheless, these minor shape variations negligibly impact entropy coding efficiency, as EG codes maintain robustness across the entire family of generalized Gaussian sources.

The sparsity of the values model parameters take is the reason that in BackSlash, we use DDGR, as opposed to EG code length directly in BackSlash - even though the same EG code may be used throughout BackSlash, the parameter value that the codeword is mapped to changes.

The mapping (termed “Value Mapping") between the quantized parameter set $Q_s$ and the codeword set $C_s$ is defined as $\mathcal{C} = \{ (q_i, c_i) \mid q_i \in Q_s, c_i \in C_s \}$.

\section{Experiments}

\begin{table*}[t]
\caption{Compression performance of BackSlash with different model architectures and parameter scales.}
\label{table:model}
\vskip 0.1in
\begin{center}
\begin{small}
\begin{tabular}{cccccc|ccc}
\toprule
\textbf{Model} & \textbf{Param Size} & \textbf{Method} & \textbf{FL (bits)} & \textbf{EG (bits)} & \textbf{HM} (bits) & \textbf{EG Compress} & \textbf{HM Compress} & \textbf{Accuracy}\\
\midrule
\multirow{2}{*}{BERT} & \multirow{2}{*}{110M} & - & 10.00 & 7.31 & 5.47 & 27\% & 45\% & \textbf{93.63\%} \\
 & & BackSlash & 10.00 & 2.64 & 2.42 & \textbf{74\%} & \textbf{76\%} & 92.59\% \\
 \midrule
\multirow{2}{*}{GPT} & \multirow{2}{*}{774M} & - & 11.00 & 7.78 & 5.73 & 29\% & 48\% & 85.92\% \\
 & & BackSlash & 11.00 & 2.46 & 2.25 & \textbf{78\%} & \textbf{80\%} & \textbf{88.73\%} \\
 \midrule
\multirow{2}{*}{Llama} & \multirow{2}{*}{1B} & - & 10.00 & 5.49 & 4.43 & 45\% & 56\% & 86.09\% \\
 & & BackSlash & 10.00 & 1.72 & 1.66 & \textbf{83\%} & \textbf{83\%} & \textbf{86.93\%} \\
 \midrule
\multirow{2}{*}{Gemma} & \multirow{2}{*}{2B} & - & 11.00 & 4.45 & 3.95 & 60\% & 64\% & \textbf{86.95\%} \\
 & & BackSlash & 11.00 & 1.16 & 1.15 & \textbf{89\%} & \textbf{90\%} & 85.86\% \\
\bottomrule
\end{tabular}
\end{small}
\end{center}
\vskip 0.1in
\end{table*}

We perform various classification tasks on popular LLMs including BERT, GPT, Llama, and Gemma to evaluate the performances of BackSlash by classification accuracy, and generation tasks on DeepSeek evaluated by next token accuracy. We mainly use classification tasks on BERT to analyze the effects of BackSlash when it is trained and deployed, as classification accuracy is one of the most intuitive quantitative metrics of model performance and BERT model has a better performance on classification tasks. In addition, we examined the entropy coding efficiency of EG codes as compared with Huffman (HM) coding and fixed-length (FL) coding with value mapping for all EG, HM and FL codes. 
\subsection{Performance}
Taking the sentiment analysis task of the BERT model as an example, we tested BackSlash using different Lagrange multiplier $\lambda$ settings. Fig.~\ref{fig:loss} shows how the loss changes in training under different Lagrange multipliers. As RD Costs may vary significantly with different Lagrange multipliers, for visual clarity, we used ($\mathcal{R}_{RD}=\log_{10}(\mathcal{J}-\beta\mathcal{J}_{min})$, $\beta=0.995$) as the y-axis. As can be seen from the figure, the larger the Lagrange multiplier, the steeper the curve, reflecting the fact that the Lagrange multiplier controls the training speed of the BackSlash.

Fig.~\ref{fig:shape} illustrates how the shape parameter of the parameter distribution varies during training.
For different $\lambda$ values, the shape parameter was set to an identical initial value but converged to different values, reflecting how $\lambda$ in the BackSlash led to different model distributions.

In Fig.~\ref{fig:avglength}, after quantizing the model parameters with the quantization step $2^{-8}$, We use EG code, HM code and FL code to encode the model parameters and compute the average code length respectively.
When applying EG coding and HM coding after unconstrained training, the model size was compressed to 73\% and 55\% of the size of FL coding, corresponding to 27\% and 45\% saving. Whereas BackSlash with EG coding and $\lambda$ of 1000, reduced the model size to 26\% and 24\%.
Even though Huffman coding leads to a very small gain in coding efficiency, a different Huffman table and the corresponding encoder/decoder will have to be designed and implemented for each LLM. In contrast, the same EG table could be used across models and sizes (e.g. DeepSeek 7B and 170B). 

Fig.~\ref{fig:accuracy} demonstrates the effect of BackSlash on model accuracy.
We can find that BackSlash with reasonable $\lambda$ did not have a significant effect on accuracy.
For the model with $\lambda=1000$, model performance decreased by only 0.02\% on the training set and 1.90\% on the test set as compared with normal training (i.e. $\lambda=0$). It was observed that model accuracy was not monotonic with regard to $\lambda$, i.e. there is an optimal $\lambda$ value, the setting of which is a topic under investigation. 

Fig.~\ref{fig:distribution} shows the impact of $\lambda$ on model parameter distribution. As can be seen clearly, as $\lambda$ increases, i.e. if we give more weights to the rate in the accuracy-rate trade-off, the model trained by BackSlash would become more sparse. 

\begin{table*}[ht]
\caption{Compression performance of BackSlash under different deep learning tasks.}
\label{table:task}
\vskip 0.1in
\begin{center}
\begin{small}
\begin{tabular}{cccccc|ccc}
\toprule
\textbf{Task} & \textbf{Dataset} & \textbf{Method} & \textbf{FL (bits)} & \textbf{EG (bits)} & \textbf{HM (bits)} & \textbf{EG Compress} & \textbf{HM Compress} & \textbf{Accuracy}\\
\midrule

\multirow{2}{*}{Sentiment} & \multirow{2}{*}{IMDB} & - & 10.00 & 7.31 & 5.47 & 27\% & 45\% & \textbf{93.63\%} \\
 & & BackSlash & 10.00 & 2.64 & 2.42 & \textbf{74\%} & \textbf{76\%} & 92.59\% \\
\midrule

\multirow{2}{*}{Spam} & \multirow{2}{*}{Enron-Spam} & - & 10.00 & 7.31 & 5.47 & 27\% & 45\% &\textbf{99.65\%} \\
 & & BackSlash & 10.00 & 2.42 & 2.19 & \textbf{76\%} & \textbf{78\%} & 98.96\% \\
\midrule

\multirow{2}{*}{Topic} & \multirow{2}{*}{20 Newsgroups} & - & 10.00 & 7.31 & 5.47 & 27\% & 45\% & \textbf{70.78\%} \\
 & & BackSlash & 10.00 & 3.61 & 3.18 & \textbf{64\%} & \textbf{68\%} & 69.36\% \\
\midrule

\multirow{2}{*}{Q-A} & \multirow{2}{*}{SQuAD} & - & 11.00 & 5.95 & 4.70 & 46\% & 57\% & 99.97\% \\
 & & BackSlash & 11.00 & 2.90 & 2.81 & \textbf{74\%} & \textbf{74\%} & \textbf{99.97\%} \\
\midrule

\multirow{2}{*}{Translation} & \multirow{2}{*}{WMT-19} & - & 11.00 & 6.10 & 4.70 & 45\% & 57\% & \textbf{99.96\%} \\
 & & BackSlash & 11.00 & 3.10 & 3.00 & \textbf{72\%} & \textbf{73\%} & 99.95\% \\
\bottomrule
\end{tabular}
\end{small}
\end{center}
\vskip 0.3in
\end{table*}

\begin{figure*}[ht]
    \centering
    \begin{minipage}{0.45\textwidth}
        \centering
        \includegraphics[width=\linewidth]{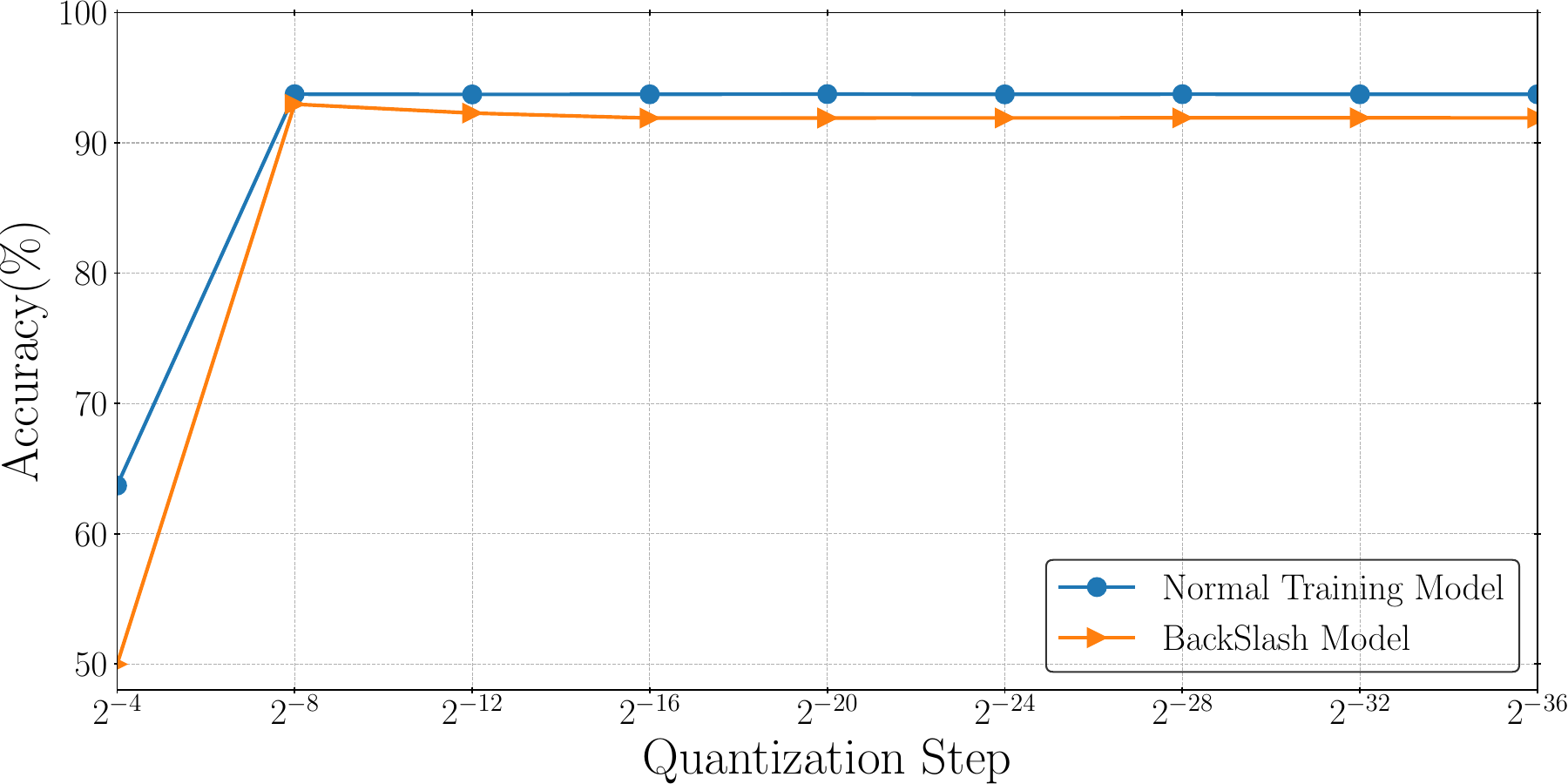}
        \vspace{-0.1in}
        \caption{Quantization using different quantization steps for BackSlash model and normal training model.}
        \vspace{0.1in}
        \label{fig:quant}
    \end{minipage}
    \hspace{0.05in}
    \begin{minipage}{0.45\textwidth}
        \centering
        \includegraphics[width=\linewidth]{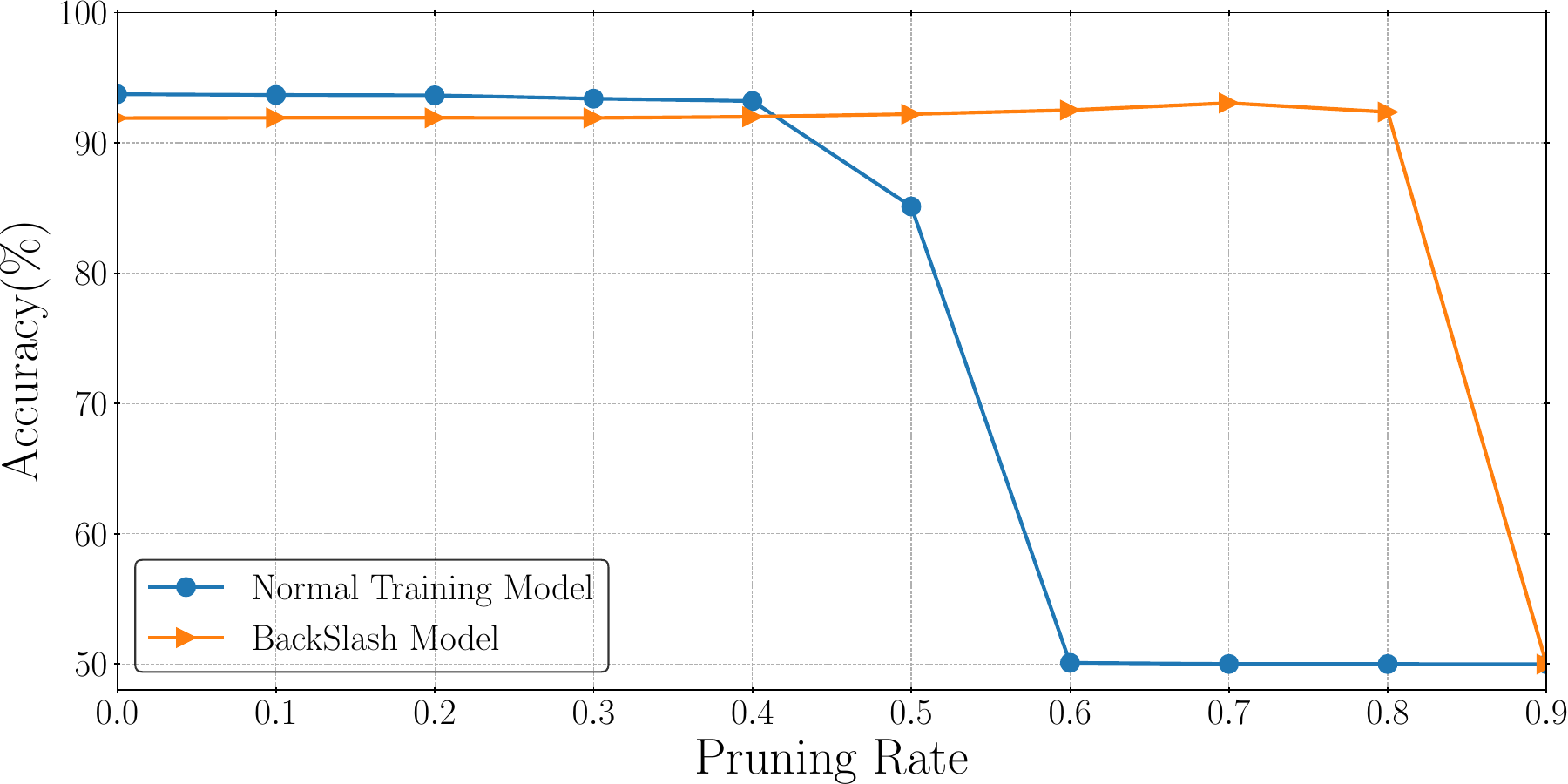}
        \vspace{-0.1in}
        \caption{Pruning using different pruning rates for BackSlash model and normal training model.}
        \vspace{0.1in}
        \label{fig:prune}
    \end{minipage}
\end{figure*}

\begin{table*}[ht]
\caption{Compression performance of BackSlash under different regularization terms.}
\label{table:regularization}
\vskip 0.1in
\begin{center}
\begin{small}
\begin{tabular}{cccccc|cc}
\toprule
\textbf{DGGG} & \textbf{Convergence Shape} & \textbf{Accuracy} & \textbf{FL (bits)} & \textbf{EG (bits)} & \textbf{HM (bits)} & \textbf{EG Compression} & \textbf{HM Compression}\\
\midrule

DGGR & 0.13 & 91.18\% & 10.00 & 1.37 & 1.32 & 86\% & 87\% \\
\midrule
$L_{0.5}$ & 0.22 & 91.88\% & 10.00 & 2.90 & 2.01 & 71\% & 80\% \\
\midrule
$L_{1}$ & 0.15 & 90.65\% & 10.00 & 1.52 & 1.46 & 85\% & 85\% \\
\midrule
$L_{2}$ & 0.10 & 88.29\% & 10.00 & 1.16 & 1.14 & 88\% & 89\% \\
\bottomrule
\end{tabular}
\end{small}
\end{center}
\vskip 0.2in
\end{table*}

In the current study, the setting of $\lambda$ was still through trials-and-errors. For example, when we set up a set of values for $\Lambda$ and train BERT model using BackSlash until convergence, we found that the model trained with $\lambda=2000$ achieved the best overall trade-off with 2.52\% loss in accuracy and only 13\% of the size. Moreover, in our extensive experiments, it consistently achieves similar and remarkable effectiveness across various models and tasks.

\subsection{Generalization Analysis}

Model architectures and training tasks are of great significance to both the process and the performance of the model and tend to affect the final model obtained from training heavily. It is worth discussing whether BackSlash has the same effects in other models and tasks besides the sentiment analysis of BERT.

In Table~\ref{table:model}, we perform the sentiment analysis task on BERT, GPT, Llama, and Gemma under normal training and BackSlash, respectively. These models are chosen because the differences in structure and parameter size among them are large enough to reflect the wide utility of BackSlash. Although different model structures introduce some variability in the results, BackSlash performs similarly for parameter compression. For all the models, BackSlash compresses them by more than 75\%, with the highest being 90\% for Gemma. Such similar performance comes from the insensitivity of BackSlash to network structure and parameter size. In addition, it can be seen that in GPT and Llama, the accuracy of using BackSlash is instead slightly higher than that of normal training, which we analyze as originating from the regularization effect attached to BackSlash.

In Table~\ref{table:task}, we perform more classification tasks on the BERT model and generation tasks on DeepSeek model under normal training and BackSlash. The “Sentiment” and “Spam” are both binary-classification tasks and the “Topic” is a 20-class-classification tasks, which are evaluated by classification accuracy. The "Q-A" and "Translation" are both text generation tasks, which are evaluated by next token accuracy. These tasks can achieve satisfactory compression performance without compromising model accuracy. BackSlash achieves approximately 70\% compression rate compared to the original size in both classification and generation tasks, demonstrating its strong generalization capability across different task types.

\subsection{Deployment}
Deployment and inference for edge devices are always the central problem and primary purpose of model compression, and quantization and pruning are the main means to deploy the fine-tuned LLMs in edge devices. Therefore, it is necessary to discuss whether the generalization ability of BackSlash models can be maintained in quantization and pruning.

Fig.~\ref{fig:quant} illustrates how the accuracy of the BERT model varies with the quantization steps under normal training and BackSlash. When the quantization step is taken $2^{-4}$, the generalization ability of both normal training and BackSlash models is completely destroyed. When the quantization step is not less than $2^{-8}$, the accuracy of both models changes very smoothly. Both models show the same trend in quantization. This is because quantization uniformly destroys the accuracy of the parameters, so whether or not to use BackSlash does not have an additional negative impact on the quantization results. Furthermore, We performed the same experiments in GPT, Llama, and Gemma, and they all showed identical results to BERT.

Fig.~\ref{fig:prune} illustrates how the accuracy of the BERT model varies with the pruning rates under normal training and BackSlash. We can see that the predictive ability of the conventionally trained model has begun to degrade when the pruning ratio reaches 50\% and has completely lost its predictive ability when it reaches 60\%. Instead, the BackSlash model continually maintains its generalization accuracy when the pruning ratio reaches 80\%. This is because BackSlash makes the model's parameter distribution more sparse, which increases the space for pruning. So BackSlash's model is also easier to deploy on edge-end devices through pruning and performs more efficient inference. Furthermore, we also performed pruning on GPT, Llama, and Gemma under BackSlash, and the maximum pruning rates for them to maintain accuracy are all 90\% while the normal training models start to lose their effectiveness at pruning rates less than 60\%, which is similar to the BERT model.

\subsection{Ablation}

As discussed in Section \ref{sec:algo}, $L_1$ and $L_2$ regularizations are special cases of DGGR assuming model parameters follow a Laplace and Gaussian Distribution, respectively. So whether such shape-specific $L_p$ regularization terms can effectively substitute DGGR in the BackSlash framework warrants further investigation.

We perform the sentiment analysis task on BERT with BackSlash using DGGR, $L_{0.5}$, $L_1$, $L_2$ respectively, to evaluate their impacts on model performance and code rate. As shown in Table~\ref{table:regularization}. Shape-specific $L_p$ regularization terms present notable theoretical and practical limitations.

From a theoretical perspective, the $L_{0.5}$, $L_1$, and $L_2$ terms implicitly assume that model parameters follow generalized Gaussian distributions with fixed shape parameters of 0.5, 1, and 2, respectively. However, the actual shape parameters of the model converge to 0.22, 0.15, and 0.10, contradicting the fixed-shape hypothesis. In contrast, DGGR's dynamic shape parameter adaptation naturally accommodates the evolving weight distribution throughout the optimization process.

From an effectiveness perspective, $L_2$ achieves marginally better compression than DGGR but incurs significant accuracy degradation, which indicating its detrimental impact on model performance. $L_1$ is inferior to DGGR in both code length and accuracy. $L_{0.5}$ demonstrates slightly better accuracy but its code length is more than twice that of DGGR, which shows its weakness in parameters compression. These findings suggest that DGGR's adaptive shape parameter adjustment puts performance and code rate in a better balance.

\section{Conclusion and Future Work}
We propose BackSlash, a training framework for LLMs that jointly optimizes model size and performance. We found that LLM parameters can be well modeled with quantized GG sources of shape parameters less than 2, and can be entropy coded with extremely high efficiency and robustness using EG codes. Experiments with popular LLMs show that BackSlash was capable of reducing model size by up to 80\% with virtually no loss in performance. 

Currently, we are conducting more experiments with more LLMs and tasks. The optimal setting of $\lambda$ is also under investigation, as well as efficient hardware architecture that can take advantage of the increased sparseness of the model in more efficient training and inference.

\section*{Impact Statement}
This paper introduces a fundamentally new approach to training large models. Instead of using standard backpropagation to train a large model and compressing it afterward, our BackSlash framework integrates efficiency directly into the training process to produce small and easy-to-deploy models. This framework can significantly influence how the next-generation foundation models are trained and deployed, both in software and hardware.

\bibliographystyle{icml2025}
\bibliography{main.bib}

\begin{thebibliography}{51}
\providecommand{\natexlab}[1]{#1}
\providecommand{\url}[1]{\texttt{#1}}
\expandafter\ifx\csname urlstyle\endcsname\relax
  \providecommand{\doi}[1]{doi: #1}\else
  \providecommand{\doi}{doi: \begingroup \urlstyle{rm}\Url}\fi

\bibitem[Bai et~al.(2023)Bai, Bai, Yang, Wang, Tan, Wang, Lin, Zhou, and
  Zhou]{Bai2023QwenVLAV}
Bai, J., Bai, S., Yang, S., Wang, S., Tan, S., Wang, P., Lin, J., Zhou, C., and
  Zhou, J.
\newblock Qwen-vl: A versatile vision-language model for understanding,
  localization, text reading, and beyond.
\newblock 2023.
\newblock URL \url{https://api.semanticscholar.org/CorpusID:261101015}.

\bibitem[Berger(2003)]{berger2003rate}
Berger, T.
\newblock Rate-distortion theory.
\newblock \emph{Wiley Encyclopedia of Telecommunications}, 2003.

\bibitem[Brand et~al.(2022)Brand, Fischer, Kopte, Windsheimer, and
  Kaup]{brand2022rdonet}
Brand, F., Fischer, K., Kopte, A., Windsheimer, M., and Kaup, A.
\newblock Rdonet: Rate-distortion optimized learned image compression with
  variable depth.
\newblock In \emph{Proceedings of the IEEE/CVF Conference on Computer Vision
  and Pattern Recognition}, pp.\  1759--1763, 2022.

\bibitem[Brown et~al.(2020)Brown, Mann, Ryder, Subbiah, Kaplan, Dhariwal,
  Neelakantan, Shyam, Sastry, Askell, Agarwal, Herbert-Voss, Krueger, Henighan,
  Child, Ramesh, Ziegler, Wu, Winter, Hesse, Chen, Sigler, teusz Litwin, Gray,
  Chess, Clark, Berner, McCandlish, Radford, Sutskever, and
  Amodei]{Brown2020LanguageMA}
Brown, T.~B., Mann, B., Ryder, N., Subbiah, M., Kaplan, J., Dhariwal, P.,
  Neelakantan, A., Shyam, P., Sastry, G., Askell, A., Agarwal, S.,
  Herbert-Voss, A., Krueger, G., Henighan, T., Child, R., Ramesh, A., Ziegler,
  D.~M., Wu, J., Winter, C., Hesse, C., Chen, M., Sigler, E., teusz Litwin, M.,
  Gray, S., Chess, B., Clark, J., Berner, C., McCandlish, S., Radford, A.,
  Sutskever, I., and Amodei, D.
\newblock Language models are few-shot learners.
\newblock \emph{ArXiv}, abs/2005.14165, 2020.
\newblock URL \url{https://api.semanticscholar.org/CorpusID:218971783}.

\bibitem[Chen et~al.(2017)Chen, Wang, and Zhang]{Chen2017DarkRankAD}
Chen, Y., Wang, N., and Zhang, Z.
\newblock Darkrank: Accelerating deep metric learning via cross sample
  similarities transfer.
\newblock \emph{ArXiv}, abs/1707.01220, 2017.
\newblock URL \url{https://api.semanticscholar.org/CorpusID:19207026}.

\bibitem[Chen et~al.(2023)Chen, Wang, Ip, and Kwong]{chen2023rate}
Chen, Y., Wang, S., Ip, H., and Kwong, S.
\newblock Rate distortion optimization with adaptive content modeling for
  random-access versatile video coding.
\newblock \emph{Information Sciences}, 645:\penalty0 119325, 2023.

\bibitem[Chiang et~al.(2023)Chiang, Shang, and Qiu]{chiang2023multi}
Chiang, J.-C., Shang, H.-Y., and Qiu, J.-J.
\newblock Multi-exposure image compression considering rate-distortion
  optimization in rendered high dynamic range image.
\newblock \emph{IEEE Open Journal of Signal Processing}, 4:\penalty0 132--147,
  2023.

\bibitem[Choi et~al.(2016)Choi, El-Khamy, and Lee]{Choi2016TowardsTL}
Choi, Y., El-Khamy, M., and Lee, J.
\newblock Towards the limit of network quantization.
\newblock \emph{ArXiv}, abs/1612.01543, 2016.
\newblock URL \url{https://api.semanticscholar.org/CorpusID:17299045}.

\bibitem[Courbariaux et~al.(2015)Courbariaux, Bengio, and
  David]{Courbariaux2015BinaryConnectTD}
Courbariaux, M., Bengio, Y., and David, J.-P.
\newblock Binaryconnect: Training deep neural networks with binary weights
  during propagations.
\newblock In \emph{Neural Information Processing Systems}, 2015.
\newblock URL \url{https://api.semanticscholar.org/CorpusID:1518846}.

\bibitem[Cover(1999)]{cover1999elements}
Cover, T.~M.
\newblock \emph{Elements of information theory}.
\newblock John Wiley \& Sons, 1999.

\bibitem[Davisson(1972)]{davisson1972rate}
Davisson, L.
\newblock Rate distortion theory: A mathematical basis for data compression.
\newblock \emph{IEEE Transactions on Communications}, 20\penalty0 (6):\penalty0
  1202--1202, 1972.

\bibitem[Fitriani et~al.(2022)Fitriani, Astuti, and Wulandari]{9742936}
Fitriani, S.~A., Astuti, Y., and Wulandari, I.~R.
\newblock Least absolute shrinkage and selection operator (lasso) and k-nearest
  neighbors (k-nn) algorithm analysis based on feature selection for diamond
  price prediction.
\newblock In \emph{2021 International Seminar on Machine Learning,
  Optimization, and Data Science (ISMODE)}, pp.\  135--139, 2022.
\newblock \doi{10.1109/ISMODE53584.2022.9742936}.

\bibitem[Fortuin et~al.(2021)Fortuin, Garriga-Alonso, Wenzel, R{\"a}tsch,
  Turner, van~der Wilk, and Aitchison]{Fortuin2021BayesianNN}
Fortuin, V., Garriga-Alonso, A., Wenzel, F., R{\"a}tsch, G., Turner, R.~E.,
  van~der Wilk, M., and Aitchison, L.
\newblock Bayesian neural network priors revisited.
\newblock \emph{ArXiv}, abs/2102.06571, 2021.
\newblock URL \url{https://api.semanticscholar.org/CorpusID:231918454}.

\bibitem[Gao et~al.(2018)Gao, Wang, and Oh]{Gao2018RateDF}
Gao, W., Wang, C., and Oh, S.
\newblock Rate distortion for model compression: From theory to practice.
\newblock In \emph{International Conference on Machine Learning}, 2018.
\newblock URL \url{https://api.semanticscholar.org/CorpusID:53111003}.

\bibitem[Glorot \& Bengio(2010)Glorot and Bengio]{glorot2010understanding}
Glorot, X. and Bengio, Y.
\newblock Understanding the difficulty of training deep feedforward neural
  networks.
\newblock In \emph{Proceedings of the thirteenth international conference on
  artificial intelligence and statistics}, pp.\  249--256. JMLR Workshop and
  Conference Proceedings, 2010.

\bibitem[Gong et~al.(2014)Gong, Liu, Yang, and Bourdev]{Gong2014CompressingDC}
Gong, Y., Liu, L., Yang, M., and Bourdev, L.~D.
\newblock Compressing deep convolutional networks using vector quantization.
\newblock \emph{ArXiv}, abs/1412.6115, 2014.
\newblock URL \url{https://api.semanticscholar.org/CorpusID:6251653}.

\bibitem[Guo et~al.(2023)Guo, Zhu, Ye, Luo, and Yang]{guo2023pre}
Guo, H., Zhu, C., Ye, M., Luo, L., and Yang, X.
\newblock Pre-encoding based temporal dependent rate--distortion optimization
  for hevc.
\newblock \emph{Signal Processing: Image Communication}, 115:\penalty0 116957,
  2023.

\bibitem[Han et~al.(2015{\natexlab{a}})Han, Mao, and Dally]{Han2015DeepCC}
Han, S., Mao, H., and Dally, W.~J.
\newblock Deep compression: Compressing deep neural network with pruning,
  trained quantization and huffman coding.
\newblock \emph{arXiv: Computer Vision and Pattern Recognition},
  2015{\natexlab{a}}.
\newblock URL \url{https://api.semanticscholar.org/CorpusID:2134321}.

\bibitem[Han et~al.(2015{\natexlab{b}})Han, Pool, Tran, and
  Dally]{Han2015LearningBW}
Han, S., Pool, J., Tran, J., and Dally, W.~J.
\newblock Learning both weights and connections for efficient neural network.
\newblock In \emph{Neural Information Processing Systems}, 2015{\natexlab{b}}.
\newblock URL \url{https://api.semanticscholar.org/CorpusID:2238772}.

\bibitem[He et~al.(2015)He, Zhang, Ren, and Sun]{He2015DelvingDI}
He, K., Zhang, X., Ren, S., and Sun, J.
\newblock Delving deep into rectifiers: Surpassing human-level performance on
  imagenet classification.
\newblock \emph{2015 IEEE International Conference on Computer Vision (ICCV)},
  pp.\  1026--1034, 2015.
\newblock URL \url{https://api.semanticscholar.org/CorpusID:13740328}.

\bibitem[He et~al.(2018)He, Lin, Liu, Wang, Li, and Han]{He2018AMCAF}
He, Y., Lin, J., Liu, Z., Wang, H., Li, L.-J., and Han, S.
\newblock Amc: Automl for model compression and acceleration on mobile devices.
\newblock In \emph{European Conference on Computer Vision}, 2018.
\newblock URL \url{https://api.semanticscholar.org/CorpusID:52048008}.

\bibitem[Hoerl \& Kennard(1970)Hoerl and Kennard]{hoerl1970ridge}
Hoerl, A.~E. and Kennard, R.~W.
\newblock Ridge regression: Biased estimation for nonorthogonal problems.
\newblock \emph{Technometrics}, 12\penalty0 (1):\penalty0 55--67, 1970.

\bibitem[Isik et~al.(2021)Isik, No, and Weissman]{Isik2021SuccessivePF}
Isik, B., No, A., and Weissman, T.
\newblock Successive pruning for model compression via rate distortion theory.
\newblock \emph{ArXiv}, abs/2102.08329, 2021.
\newblock URL \url{https://api.semanticscholar.org/CorpusID:231933836}.

\bibitem[Itu-T \& Jtc(2010)Itu-T and Jtc]{Jtc2010AdvancedVC}
Itu-T and Jtc, I.~I.
\newblock Advanced video coding for generic audiovisual services.
\newblock 2010.
\newblock URL \url{https://api.semanticscholar.org/CorpusID:60356047}.

\bibitem[Jaderberg et~al.(2014)Jaderberg, Vedaldi, and
  Zisserman]{jaderberg2014speeding}
Jaderberg, M., Vedaldi, A., and Zisserman, A.
\newblock Speeding up convolutional neural networks with low rank expansions.
\newblock \emph{arXiv preprint arXiv:1405.3866}, 2014.

\bibitem[Kossaifi et~al.(2019)Kossaifi, Bulat, Tzimiropoulos, and
  Pantic]{Kossaifi2019TNetPF}
Kossaifi, J., Bulat, A., Tzimiropoulos, G., and Pantic, M.
\newblock T-net: Parametrizing fully convolutional nets with a single
  high-order tensor.
\newblock \emph{2019 IEEE/CVF Conference on Computer Vision and Pattern
  Recognition (CVPR)}, pp.\  7814--7823, 2019.
\newblock URL \url{https://api.semanticscholar.org/CorpusID:102353394}.

\bibitem[Li \& Liu(2016)Li and Liu]{Li2016TernaryWN}
Li, F. and Liu, B.
\newblock Ternary weight networks.
\newblock \emph{ICASSP 2023 - 2023 IEEE International Conference on Acoustics,
  Speech and Signal Processing (ICASSP)}, pp.\  1--5, 2016.
\newblock URL \url{https://api.semanticscholar.org/CorpusID:13556195}.

\bibitem[Li et~al.(2016)Li, Kadav, Durdanovic, Samet, and
  Graf]{Li2016PruningFF}
Li, H., Kadav, A., Durdanovic, I., Samet, H., and Graf, H.~P.
\newblock Pruning filters for efficient convnets.
\newblock \emph{ArXiv}, abs/1608.08710, 2016.
\newblock URL \url{https://api.semanticscholar.org/CorpusID:14089312}.

\bibitem[Lin et~al.(2017)Lin, Han, Mao, Wang, and Dally]{Lin2017DeepGC}
Lin, Y., Han, S., Mao, H., Wang, Y., and Dally, W.~J.
\newblock Deep gradient compression: Reducing the communication bandwidth for
  distributed training.
\newblock \emph{ArXiv}, abs/1712.01887, 2017.
\newblock URL \url{https://api.semanticscholar.org/CorpusID:38796293}.

\bibitem[Liu et~al.(2021)Liu, Cheng, Huang, Xing, and
  Shen]{Liu2021NonuniformtoUniformQT}
Liu, Z., Cheng, K.-T., Huang, D., Xing, E.~P., and Shen, Z.
\newblock Nonuniform-to-uniform quantization: Towards accurate quantization via
  generalized straight-through estimation.
\newblock \emph{2022 IEEE/CVF Conference on Computer Vision and Pattern
  Recognition (CVPR)}, pp.\  4932--4942, 2021.
\newblock URL \url{https://api.semanticscholar.org/CorpusID:244715141}.

\bibitem[Long et~al.(2020)Long, Zeng, Ben, Zhou, and Zhang]{long2020novel}
Long, X., Zeng, X., Ben, Z., Zhou, D., and Zhang, M.
\newblock A novel low-bit quantization strategy for compressing deep neural
  networks.
\newblock \emph{Computational Intelligence and Neuroscience}, 2020\penalty0
  (1):\penalty0 7839064, 2020.

\bibitem[Luo et~al.(2017)Luo, Wu, and Lin]{Luo2017ThiNetAF}
Luo, J.-H., Wu, J., and Lin, W.
\newblock Thinet: A filter level pruning method for deep neural network
  compression.
\newblock \emph{2017 IEEE International Conference on Computer Vision (ICCV)},
  pp.\  5068--5076, 2017.
\newblock URL \url{https://api.semanticscholar.org/CorpusID:11169209}.

\bibitem[Luttrell et~al.(2000)Luttrell, Wen, and
  Villasenor]{luttrell2000trellis}
Luttrell, M., Wen, J., and Villasenor, J.~D.
\newblock Trellis-based rd optimal quantization in h. 263+.
\newblock In \emph{Proceedings 2000 International Conference on Image
  Processing (Cat. No. 00CH37101)}, volume~2, pp.\  852--854. IEEE, 2000.

\bibitem[Masana et~al.(2017)Masana, van~de Weijer, Herranz, Bagdanov, and
  {\'A}lvarez]{Masana2017DomainAdaptiveDN}
Masana, M., van~de Weijer, J., Herranz, L., Bagdanov, A.~D., and {\'A}lvarez,
  J.~M.
\newblock Domain-adaptive deep network compression.
\newblock \emph{2017 IEEE International Conference on Computer Vision (ICCV)},
  pp.\  4299--4307, 2017.
\newblock URL \url{https://api.semanticscholar.org/CorpusID:11067299}.

\bibitem[Park et~al.(2023)Park, Kim, Kim, Choi, and Lee]{Park2023DynamicSP}
Park, J.-H., Kim, Y., Kim, J., Choi, J.-Y., and Lee, S.
\newblock Dynamic structure pruning for compressing cnns.
\newblock \emph{ArXiv}, abs/2303.09736, 2023.
\newblock URL \url{https://api.semanticscholar.org/CorpusID:257622926}.

\bibitem[Rastegari et~al.(2016)Rastegari, Ordonez, Redmon, and
  Farhadi]{Rastegari2016XNORNetIC}
Rastegari, M., Ordonez, V., Redmon, J., and Farhadi, A.
\newblock Xnor-net: Imagenet classification using binary convolutional neural
  networks.
\newblock \emph{ArXiv}, abs/1603.05279, 2016.
\newblock URL \url{https://api.semanticscholar.org/CorpusID:14925907}.

\bibitem[Shannon(1948)]{shannon1948mathematical}
Shannon, C.~E.
\newblock A mathematical theory of communication.
\newblock \emph{The Bell system technical journal}, 27\penalty0 (3):\penalty0
  379--423, 1948.

\bibitem[Sharifi \& Leon-Garcia(1995)Sharifi and
  Leon-Garcia]{Sharifi1995EstimationOS}
Sharifi, K. and Leon-Garcia, A.
\newblock Estimation of shape parameter for generalized gaussian distributions
  in subband decompositions of video.
\newblock \emph{IEEE Trans. Circuits Syst. Video Technol.}, 5:\penalty0 52--56,
  1995.
\newblock URL \url{https://api.semanticscholar.org/CorpusID:41130607}.

\bibitem[Touvron et~al.(2023)Touvron, Lavril, Izacard, Martinet, Lachaux,
  Lacroix, Rozi{\`e}re, Goyal, Hambro, Azhar, Rodriguez, Joulin, Grave, and
  Lample]{Touvron2023LLaMAOA}
Touvron, H., Lavril, T., Izacard, G., Martinet, X., Lachaux, M.-A., Lacroix,
  T., Rozi{\`e}re, B., Goyal, N., Hambro, E., Azhar, F., Rodriguez, A., Joulin,
  A., Grave, E., and Lample, G.
\newblock Llama: Open and efficient foundation language models.
\newblock \emph{ArXiv}, abs/2302.13971, 2023.
\newblock URL \url{https://api.semanticscholar.org/CorpusID:257219404}.

\bibitem[Wang et~al.(2018{\natexlab{a}})Wang, Liu, Lin, Lin, and
  Han]{Wang2018HAQHA}
Wang, K., Liu, Z., Lin, Y., Lin, J., and Han, S.
\newblock Haq: Hardware-aware automated quantization with mixed precision.
\newblock \emph{2019 IEEE/CVF Conference on Computer Vision and Pattern
  Recognition (CVPR)}, pp.\  8604--8612, 2018{\natexlab{a}}.
\newblock URL \url{https://api.semanticscholar.org/CorpusID:102350477}.

\bibitem[Wang et~al.(2018{\natexlab{b}})Wang, Hu, Zhang, Zhang, Liu, and
  Cheng]{8578558}
Wang, P., Hu, Q., Zhang, Y., Zhang, C., Liu, Y., and Cheng, J.
\newblock Two-step quantization for low-bit neural networks.
\newblock In \emph{2018 IEEE/CVF Conference on Computer Vision and Pattern
  Recognition}, pp.\  4376--4384, 2018{\natexlab{b}}.
\newblock \doi{10.1109/CVPR.2018.00460}.

\bibitem[Wang et~al.(2017)Wang, Xu, Xu, and Tao]{Wang2017BeyondFC}
Wang, Y., Xu, C., Xu, C., and Tao, D.
\newblock Beyond filters: Compact feature map for portable deep model.
\newblock In \emph{International Conference on Machine Learning}, 2017.
\newblock URL \url{https://api.semanticscholar.org/CorpusID:29145201}.

\bibitem[Wen \& Villasenor(1999)Wen and Villasenor]{761289}
Wen, J. and Villasenor, J.
\newblock Structured prefix codes for quantized low-shape-parameter generalized
  gaussian sources.
\newblock \emph{IEEE Transactions on Information Theory}, 45\penalty0
  (4):\penalty0 1307--1314, 1999.
\newblock \doi{10.1109/18.761289}.

\bibitem[Wien(2015)]{wien2015high}
Wien, M.
\newblock High efficiency video coding.
\newblock \emph{Coding Tools and specification}, 24:\penalty0 1, 2015.

\bibitem[Xia et~al.(2023{\natexlab{a}})Xia, Tsang, and
  Lau]{Xia2023StructuredBC}
Xia, C.-G., Tsang, D. H.-K., and Lau, V. K.~N.
\newblock Structured bayesian compression for deep neural networks based on the
  turbo-vbi approach.
\newblock \emph{IEEE Transactions on Signal Processing}, 71:\penalty0 670--685,
  2023{\natexlab{a}}.
\newblock URL \url{https://api.semanticscholar.org/CorpusID:257050720}.

\bibitem[Xia et~al.(2023{\natexlab{b}})Xia, Jin, Meng, Ding, and
  Zhang]{xia2023gan}
Xia, F., Jin, J., Meng, L., Ding, F., and Zhang, H.
\newblock Gan-based image compression with improved rdo process.
\newblock In \emph{International Conference on Image and Graphics}, pp.\
  361--372. Springer, 2023{\natexlab{b}}.

\bibitem[Xu et~al.(2018)Xu, Ouyang, Wang, and Sebe]{Xu2018PADNetMG}
Xu, D., Ouyang, W., Wang, X., and Sebe, N.
\newblock Pad-net: Multi-tasks guided prediction-and-distillation network for
  simultaneous depth estimation and scene parsing.
\newblock \emph{2018 IEEE/CVF Conference on Computer Vision and Pattern
  Recognition}, pp.\  675--684, 2018.
\newblock URL \url{https://api.semanticscholar.org/CorpusID:21670200}.

\bibitem[Zhai et~al.(2023)Zhai, Guo, Liu, Xing, and Xu]{Zhai2023LAPPLA}
Zhai, P., Guo, K., Liu, F., Xing, X., and Xu, X.
\newblock Lapp: Layer adaptive progressive pruning for compressing cnns from
  scratch.
\newblock \emph{ArXiv}, abs/2309.14157, 2023.
\newblock URL \url{https://api.semanticscholar.org/CorpusID:262459258}.

\bibitem[Zhang et~al.(2024)Zhang, Lu, Liang, Tang, Hu, and
  Song]{zhang2024efficient}
Zhang, Z., Lu, G., Liang, H., Tang, A., Hu, Q., and Song, L.
\newblock Efficient dynamic-nerf based volumetric video coding with rate
  distortion optimization.
\newblock \emph{arXiv preprint arXiv:2402.01380}, 2024.

\bibitem[Zhou et~al.(2016)Zhou, Ni, Zhou, Wen, Wu, and
  Zou]{Zhou2016DoReFaNetTL}
Zhou, S., Ni, Z., Zhou, X., Wen, H., Wu, Y., and Zou, Y.
\newblock Dorefa-net: Training low bitwidth convolutional neural networks with
  low bitwidth gradients.
\newblock \emph{ArXiv}, abs/1606.06160, 2016.
\newblock URL \url{https://api.semanticscholar.org/CorpusID:14395129}.

\bibitem[Zhu et~al.(2016)Zhu, Han, Mao, and Dally]{Zhu2016TrainedTQ}
Zhu, C., Han, S., Mao, H., and Dally, W.~J.
\newblock Trained ternary quantization.
\newblock \emph{ArXiv}, abs/1612.01064, 2016.
\newblock URL \url{https://api.semanticscholar.org/CorpusID:224893}.

\end{thebibliography}

\end{document}